\documentclass[12pt]{article}

\usepackage{sbc-template}

\usepackage{graphicx,url}

\usepackage{graphicx,latexsym,graphics}
\usepackage{algorithm}
\usepackage{algorithmic}
\usepackage{subfigure}
\usepackage{amsmath}
\usepackage{amssymb}
\usepackage{graphicx,color}

\title{Fourier Analysis and $q$-Gaussian Functions: Analytical
and Numerical Results}

\author{Paulo S. S. Rodrigues\inst{1} and Gilson A. Giraldi\inst{2}}

\address{Departamento de Engenharia El\'{e}trica \\
  FEI, S\~{a}o Bernardo do Campo, SP, Brazil
\nextinstitute
Laborat\'{o}rio Nacional de Computa\c c\~ao Cient\'{i}fica \\
  Petrop\'{o}lis, RJ, Brazil
  \email{psergio@fei.edu.br, gilson@lncc.br}
}

\begin{document}

\maketitle

\begin{abstract}
It is a consensus in signal processing that the Gaussian kernel and
its partial derivatives enable the development of robust algorithms
for feature detection. Fourier analysis and convolution theory have
central role in such development. In this paper we collect theoretical
elements to follow this avenue but using the $q$-Gaussian kernel that
is a nonextensive generalization of the Gaussian one. Firstly, we
review some theoretical elements behind the one-dimensional $q$-Gaussian
and its Fourier transform. Then, we consider the two-dimensional $q$-Gaussian
and we highlight the issues behind its analytical Fourier transform
computation. We analyze the $q$-Gaussian kernel in the space and Fourier
domains using the concepts of space window, cut-off frequency, and
the Heisenberg inequality.
\end{abstract}


\section{Introduction}

Feature extraction is an essential step for image analysis and computer
vision tasks such as image matching and object recognition \cite{Jain89}.
The specific case of edge detection, has been extensively considered
in the image processing literature \cite{Gonzalez1992}.
In this subject, the Gaussian kernel and its partial derivatives have
inspired a wide range of works in the image analysis literature for
the development of multiscale approaches \cite{LopezMolina2013}.

These works had established the background for multiscale representation
based on the viewpoint of the functional structure of digital images
\cite{Florack92,Koenderink84}. Basically, the grayscale
of the observed image is realized as a general function $f$ of the
space of square integrable functions on $\mathbb{R}^{2}$, denoted
by $L^{2}\left(\mathbb{R}^{2}\right)$. The linear scale-space is
generated by the convolution with scaled Gaussian kernels with filtering
properties analyzed in the frequency space given by the Fourier transform.
Such approach can be seen from the isotropic diffusion equation viewpoint,
which opens the possibility of generating more general multiscale
representations based on the anisotropic diffusion equation \cite{Perona90}.

The Gaussian kernel plays also a fundamental role in statistical physics
due to the fact that an enormous amount of phenomena in nature follow
the Gaussian distribution and the extensive thermostatistics governed
by the Boltzmann-Gibbs entropy and the standard central limit theorem
\cite{Liboff1990}. More recently, Tsallis nonextensive entropy and
generalizations of the central limit theorem gives the foundations
for noextensive counterparts of the Boltzmann-Gibbs statistical mechanics
\cite{tsallisbooka,Tsallis1988}. In this context, generalization
of Gaussian smoothing kernel within the Tsallis nonextensive scenario,
named $q$-Gaussian, has been proposed \cite{Tsallis1995}. Also, the
difference of $q$-Gaussian kernels is introduced in \cite{Assirati-Bruno2014}
to build a method for edge extraction in digital images. These works
demonstrate the potential of the $q$-Gaussian based methods by comparing
the obtained results with the ones obtained with traditional techniques
that rely on the Gaussian function \cite{Gallao-Rodrigues2015}. However,
the behaviour of the $q$-Gaussian in the frequency domain has been ignored.

In this paper, we collect theoretical elements, published in the references
\cite{Borges-Tsallis-Miranda:2004,Borges0305:1998,Daaz20091}, to perform such analysis. We review
the $q$-exponential function and the $q$-Gaussian distribution. Then,
we offer details of the Fourier transform computation for $d=1$.
The two-dimensional $q$-Gaussian is also considered from the analytical
and numerical viewpoint. In the experimental results, we analyze the
$q$-Gaussian in the frequency domain and compare its profile with the
Gaussian one. In fact, we consider the Fourier transform of the one-dimensional
$q$-Gaussian which emphasizes the fact that a $q$-Gaussian is a low-pass
filter. We study the size of the space window and analyze the influence of the parameter $q$
in the cut-off frequency and in the Heisenberg inequality.


The paper is organized as follows. Section \ref{sec:Tsallis-Entropy-and}
focuses on Tsallis entropy, the $q$-exponential and $q$-Gaussian and summarizes
some of their basic properties. The Fourier transform computation
of the $q$-Gaussian is discussed on sections \ref{sec:Fourier-Transform-1D}
and \ref{sec:Fourier-Transform-2D}. The computational results are
presented on section \ref{sec:Computational-Experiments}. The Appendices
\ref{sec:Appendix-A-Gaussian}-\ref{sec:Appendix-F-Special-Functions}
complete the material with details about the $q$-Gaussian and its Fourier
transform as well as the special functions Gamma, Whittaker, Bessel
and Beta functions, used along the text.

\section{Tsallis Entropy and $q$-Gaussian\label{sec:Tsallis-Entropy-and}}

In the last decade, Tsallis \cite{Tsallis1988} has proposed the
following generalized nonextensive entropic form:
\begin{equation}
S_{q}=k\frac{1-\sum_{i=1}^{W}p_{i}^{q}}{q-1},\label{eq:tsallis01}
\end{equation}
where $k$ is a positive constant, $p_{i}$ is a probability distribution
and $q\in\Re$ is called the entropic index. This expression recovers
the Shannon entropy in the limit $q\rightarrow1$. The Tsallis entropy
offers a new formalism in which the real parameter $q$ quantifies
the level of nonextensivity of a physical systems \cite{tsallis1999}.
In particular a general principle of maximum entropy (PME) has been
considered to find out the distribution $p_{i}$ to describe such
systems. In this PME, the goal is to find the maximum of $S_{q}$
subjected to:
\begin{equation}
\sum_{i=1}^{W}p_{i}=1,\label{prob00}
\end{equation}

\begin{equation}
\frac{\sum_{i=1}^{W}e_{i}p_{i}^{q}}{\sum_{i=1}^{W}p_{i}^{q}}=U_{q},\label{qmean}
\end{equation}
where $U_{q}$ is a known application dependent value and $e_{i}$
represent the possible states of the system (in image processing,
the gray-level intensities). Expression (\ref{prob00}) just imposes
that $p_{i}$ is probability and equation (\ref{qmean}) is a generalized
expectation value of the $e_{i}$ (if $q=1$ we get the usual mean
vale). The proposed PME can be solved using Lagrange multipliers and
the solution has the form \cite{tsallis1999,Tsallis-RenioS-Plastino:1998}:

\begin{equation}
p_{j}=\frac{\left[1-(1-q)\widetilde{\beta}e_{j}\right]^{\frac{1}{1-q}}}{\widetilde{Z}_{q}},\label{eq:prob-final}
\end{equation}
where $\widetilde{\beta}$ and $\widetilde{Z}_{q}$ are defined by the
expressions:

\[
\widetilde{\beta}=\frac{\beta}{\sum_{j=1}^{W}p_{j}^{q}+\left(1-q\right)\beta U_{q}},
\]

\[
\widetilde{Z}_{q}=\sum_{j=1}^{W}\left[1-(1-q)\widetilde{\beta}e_{j}\right]^{\frac{1}{1-q}},
\]
with $\beta$ being the Lagrange multiplier associated with the constraint
given by expression (\ref{qmean}) and, if $q<1$, then $p_{i}=0$
whenever $1-(1-q)\widetilde{\beta}e_{j}<0$ (cut-off condition). The
expressions (\ref{eq:tsallis01}) and (\ref{eq:prob-final}) inspire
the definition of the $q$-exponential function \cite{Tsallis-Nova1994}:

\begin{align}
\exp_{q}\left(x\right)=\begin{cases}
\left[1+\left(1-q\right)x\right]^{\frac{1}{1-q}}, & \text{if} \;1+\left(1-q\right)x>0,\\
0, & \text{otherwise}.
\end{cases}\label{eq:group-springs-02}
\end{align}


It can be shown that the traditional exponential function ($\exp$)
is given by the limit:

\begin{equation}
\exp\left(x\right)=\lim_{q\rightarrow1}\exp_{q}\left(x\right).\label{eq:limite-q-exponential}
\end{equation}

The equation (\ref{eq:limite-q-exponential}) motivates the definition
of the $d$-dimensional $q$-Gaussian as \cite{Daaz20091}:

\begin{equation}
G_{d,q}\left(\mathbf{x},\Sigma,\beta\right)=C_{d,q}\left(\Sigma,\beta\right)\exp_{q}\left(-\beta\mathbf{x^{T}}\Sigma^{-1}\mathbf{x}\right),\label{eq:Q-Gaussian-d}
\end{equation}
where:

\begin{equation}
C_{d,q}\left(\Sigma,\beta\right)=\left(\int_{\mathbb{R}^{d}}\exp_{q}\left(-\beta\mathbf{x^{T}}\Sigma^{-1}\mathbf{x}\right)d\mathbf{x}\right)^{-1}.\label{eq:normalization-factor-Q-Gaussian}
\end{equation}
and $\Sigma$ is the covariance matrix (symmetric and positive definite).
Due to expression (\ref{eq:limite-q-exponential}) it is straightforward
to show that:

\begin{equation}
\lim_{q\rightarrow1}\exp_{q}\left(-\beta\mathbf{x^{T}}\Sigma^{-1}\mathbf{x}\right)=\exp\left(-\beta\mathbf{x^{T}}\Sigma^{-1}\mathbf{x}\right).\label{eq:limite-d-dimensional-q-gauss00}
\end{equation}

Consequently, depending on the $C_{d,q}\left(\beta\right)$ functional
form, we can recover the $d$-dimensional Gaussian:

\begin{equation}
G_{d}\left(\mathbf{x},\Sigma\right)=\left[\left(2\pi\right)^{d/2}|\Sigma|^{1/2}\right]^{-1}\exp\left(-\frac{1}{2}\mathbf{x^{T}}\Sigma^{-1}\mathbf{x}\right),\label{eq:gaussian-2d-1}
\end{equation}
by taking the limit of $G_{d,q}\left(\mathbf{x},\Sigma,\beta\right)$at
$q\rightarrow1$. In the Appendices \ref{sec:Appendix-C-One-Dimensional} and \ref{sec:Appenix-D-Two-Dimensional} we give details of
that developments showing that, if $\beta=1/2$, then we obtain the
one and two-dimensional Gaussian functions in this way. The corresponding
$q$-expressions are reproduced bellow.

a) One-Dimensional $q$-Gaussian (see Appendix \ref{sec:Appendix-C-One-Dimensional}):

\begin{equation}
G_{1,q}\left(x,\sigma,\beta\right)=C_{1,q}\left(\sigma,\beta\right)\exp_{q}\left(-\frac{\beta}{\sigma^{2}}x^{2}\right)=C_{1,q}\left(\beta\right)\left[1+\left(q-1\right)\frac{\beta}{\sigma^{2}}x^{2}\right]^{\frac{1}{1-q}},\label{eq:Q-Gaussian-d-1-1}
\end{equation}
where:

\begin{equation}
C_{1,q}\left(\sigma,\beta\right)=\frac{\Gamma\left(\frac{1}{q-1}\right)\left[\left(q-1\right)\frac{\beta}{\sigma^{2}}\right]^{1/2}}{\sqrt{\pi}\Gamma\left(\frac{1}{q-1}-\frac{1}{2}\right)},\quad3>q>1,\label{eq:normalization-constant-q-gaussian00}
\end{equation}

\begin{equation}
C_{1,q}\left(\sigma,\beta\right)=\frac{\Gamma\left(\frac{1}{1-q}+\frac{3}{2}\right)\left(\left(1-q\right)\frac{\beta}{\sigma^{2}}\right)^{1/2}}{\sqrt{\pi}\Gamma\left(\frac{1}{1-q}+1\right)},\quad q<1,\quad|\mathbf{x}|\leq\left(\left(1-q\right)\frac{\beta}{\sigma^{2}}\right)^{-1/2}.\label{eq:normalization-constant-q-gaussian01}
\end{equation}

b) Two-Dimensional $q$-Gaussian (see Appendix \ref{sec:Appenix-D-Two-Dimensional}):

\begin{equation}
G_{2,q}\left(\mathbf{x},\Sigma,\beta\right)=\frac{\beta\left(2-q\right)}{\pi\sqrt{|\Sigma|}}\left[1+\left(1-q\right)\left(-\beta\mathbf{x^{T}}\Sigma^{-1}\mathbf{x}\right)\right]^{^{\frac{1}{1-q}}},\quad1<q<2,\label{eq:final-2D-gaussian00-1}
\end{equation}

\begin{equation}
G_{2,q}\left(\mathbf{x},\Sigma,\beta\right)=\frac{\beta\left(2-q\right)}{\pi\sqrt{|\Sigma|}}\left[1+\left(1-q\right)\left(-\beta\mathbf{x^{T}}\Sigma^{-1}\mathbf{x}\right)\right]^{^{\frac{1}{1-q}}},\quad q<1,\label{eq:final-2D-gaussian01-1}
\end{equation}
subject to the constraint:
\begin{equation}
0<\left(\mathbf{x}^{T}\Sigma^{-1}\mathbf{x}\right)^{1/2}<\frac{1}{\sqrt{\beta\left(1-q\right)}}.
\end{equation}

\section{Fourier Transform of 1D $q$-Gaussian\label{sec:Fourier-Transform-1D}}

From expression (\ref{eq:Q-Gaussian-d-1-1}) we can compute the Fourier
transform of the one-dimensional $q$-Gaussian as:

\[
\mathcal{F}\left(C_{1,q}\left(\sigma,\beta\right)\exp_{q}\left(-\frac{\beta}{\sigma^{2}}x^{2}\right);y\right)
\]

\[
=C_{1,q}\left(\sigma,\beta\right)\mathcal{F}\left(\exp_{q}\left(-\frac{\beta}{\sigma^{2}}x^{2}\right);y\right),
\]
where $\mathcal{F}\left(g\left(x\right);y\right)$means the Fourier
transform of function $g$ computed at the frequency $y$. In this
paper, the Fourier transform is defined by:

\begin{equation}
\mathcal{F}\left(g\left(x\right);y\right)=\int_{-\infty}^{+\infty}\exp\left(-2j\pi xy\right)g\left(x\right)dx.\label{eq:FT-Definition}
\end{equation}

The Appendix \ref{sec:Appendix-E-Fourier} develops the computation of the Fourier transform
for a $q$-exponential. So, using expressions (\ref{eq:fourier-1d-transform-q-gauss00})
and (\ref{eq:fourier-1d-transform-q-gauss01}) we obtain:

\[
\mathcal{F}\left(G_{1,q}\left(x,\sigma,\beta\right);y\right)=C_{1,q}\left(\sigma,\beta\right)\left[\left(q-1\right)\frac{\beta}{\sigma^{2}}\right]^{-1/2}\times
\]

\[
\left(-sign\left(2\pi\left[\left(q-1\right)\frac{\beta}{\sigma^{2}}\right]^{-1/2}y\right)2\pi\left(2^{\frac{1}{1-q}}\right)\frac{|2\pi\left[\left(q-1\right)\frac{\beta}{\sigma^{2}}\right]^{-1/2}y|^{\frac{1}{q-1}-1}}{\Gamma\left(\frac{1}{q-1}\right)}\right)\times
\]

\begin{equation}
W_{0,\frac{1}{2}+\frac{1}{1-q}}\left(2|2\pi\left[\left(q-1\right)\frac{\beta}{\sigma^{2}}\right]^{-1/2}y|\right),\;1<q<3,\label{eq:FT-of-q-gaussian-q-maior-1}
\end{equation}
where $W_{0,\frac{1}{2}+\frac{1}{1-q}}$ are Whittaker functions,
defined in the Appendix \ref{sec:Appendix-F-Special-Functions}. Consequently :

\[
abs\left[\mathcal{F}\left(G_{1,q}\left(x,\sigma,\beta\right);y\right)\right]=abs\left[C_{1,q}\left(\beta\right)\right]\times
\]

\[
abs\left[\pi^{3/2}\frac{\left[\left(q-1\right)a\right]^{-1/2}}{\Gamma\left(\frac{1}{q-1}\right)}\left(i\exp\left(\frac{i\pi}{2}\left(-\frac{1}{2}-\frac{1}{1-q}\right)\right)\right)\right]\times
\]

\begin{equation}
abs\left[\left(\frac{z}{2}\right)^{\frac{1}{q-1}-\frac{1}{2}}\mathbb{J}_{-\frac{1}{2}+\frac{1}{q-1}}\left(iz\right)+i\frac{\left[\left(\frac{z}{2}\right)^{\frac{1}{q-1}-\frac{1}{2}}\mathbb{J}_{-\frac{1}{2}+\frac{1}{q-1}}\left(iz\right)\cos\left(\pi\left(-\frac{1}{2}+\frac{1}{q-1}\right)\right)-\left(i\right)^{\frac{1}{2}-\frac{1}{q-1}}S\left(z\right)\right]}{\sin\left(\pi\left(-\frac{1}{2}+\frac{1}{q-1}\right)\right)}\right],\label{eq:abs-FT-qGaussian-q-maior-1}
\end{equation}
where $z=|2\pi\left[\left(q-1\right)a\right]^{-1/2}y|$, $a=\frac{\beta}{\sigma^{2}}$
and $1<q<3$ and:

\[
S\left(z\right)=\sum_{k=0}^{+\infty}\frac{\left(\frac{z^{2}}{4}\right)^{k}}{k!\Gamma\left(\frac{1}{2}-\frac{1}{q-1}+k+1\right)}
\]

Analogously, we can show that:

\[
\mathcal{F}\left(G_{1,q}\left(x,\sigma,\beta\right);y\right)=C_{1,q}\left(\sigma,\beta\right)\frac{\sqrt{\pi}}{\left(\left(1-q\right)\frac{\beta}{\sigma^{2}}\right)^{1/2}}\times
\]

\begin{equation}
\left(-\frac{\left(\left(1-q\right)\frac{\beta}{\sigma^{2}}\right)^{1/2}}{\pi y}\right)^{\frac{1}{1-q}+\frac{1}{2}}\Gamma\left(\frac{1}{1-q}+1\right)\mathbb{J}_{\frac{1}{1-q}+\frac{1}{2}}\left(-\frac{2\pi y}{\left(\left(1-q\right)\frac{\beta}{\sigma^{2}}\right)^{1/2}}\right),\quad q<1,\; y\neq0,\label{eq:FT-of-q-gaussian-q-menor-1}
\end{equation}

\[
\mathcal{F}\left(G_{1,q}\left(x,\sigma,\beta\right);0\right)
\]

\begin{equation}
=C_{1,q}\left(\sigma,\beta\right)\frac{2^{\frac{2}{1-q}+1}}{\left(\left(1-q\right)\frac{\beta}{\sigma^{2}}\right)^{1/2}}\frac{\Gamma\left(\frac{1}{\left(1-q\right)}+1\right)\Gamma\left(\frac{1}{\left(1-q\right)}+1\right)}{\Gamma\left(\frac{2}{1-q}+2\right)},\quad q<1,\; y=0,\label{eq:FT-of-q-gaussian-q-menor-1-y-0}
\end{equation}
with $\mathbb{J}_{\frac{1}{1-q}+\frac{1}{2}}$ being the Bessel functions
(see Appendix \ref{sec:Appendix-F-Special-Functions}).

\section{Fourier Transform of 2D $q$-Gaussian \label{sec:Fourier-Transform-2D} }

In this section we offer a sketch of the Fourier transform for the
two-dimensional $q$-Gaussian, defined by expressions (\ref{eq:final-2D-gaussian00-1})-(\ref{eq:final-2D-gaussian01-1}),
for a diagonal matrix $\Sigma=diag\left(\begin{array}{cc}
\sigma_{1}^{2} & \sigma_{2}^{2}\end{array}\right)$ which is computed as follows:

\begin{equation}
\mathcal{F}\left(G_{2,q}\left(\mathbf{x},\Sigma,\beta\right);\omega_{1},\omega_{2}\right)= C_{2,q} \times
\end{equation}

\begin{equation}
=\left(\Sigma,\beta\right)\int_{-\infty}^{+\infty}\int_{-\infty}^{+\infty}\exp\left[-2\pi j\left(\omega_{1}x+\omega_{2}y\right)\right]\left[1+\left(1-q\right)\left(-\beta\left(\frac{x^{2}}{\sigma_{1}^{2}}+\frac{y^{2}}{\sigma_{2}^{2}}\right)\right)\right]^{^{\frac{1}{1-q}}}dxdy,\label{eq:2D-QGaussian-FT00}
\end{equation}
where $\mathbf{x}=\left(x,y\right)$. Therefore, we can re-write
this expression as:

\begin{equation}
\mathcal{F}\left(G_{2,q}\left(\mathbf{z},\Sigma,\beta\right);\omega_{1},\omega_{2}\right)=C_{2,q}\left(\Sigma,\beta\right)\sqrt{|\Sigma|}\times
\end{equation}

\begin{equation}
\int_{-\infty}^{+\infty}\int_{-\infty}^{+\infty}\exp\left[-2\pi j\left(\omega_{1}z_{1}+\omega_{2}z_{2}\right)\right]\left[1+\left(1-q\right)\left(-\beta\left(z_{1}^{2}+z_{2}^{2}\right)\right)\right]^{^{\frac{1}{1-q}}}dz_{1}dz_{2}\label{eq:2D-QGaussian-FT01}
\end{equation}
where $z_{1}=x/\sigma_{1}$ and $z_{2}=x/\sigma_{2}$.

Considering just the double integral of equation (\ref{eq:2D-QGaussian-FT01})
we can write:

\[
\int_{-\infty}^{+\infty}\exp\left[-2\pi j\omega_{2}z_{2}\right]\left\{ \int_{-\infty}^{+\infty}\exp\left[-2\pi j\omega_{1}z_{1}\right]\left[1+\left(1-q\right)\left(-\beta\left(z_{1}^{2}+z_{2}^{2}\right)\right)\right]^{^{\frac{1}{1-q}}}dz_{1}\right\} dz_{2}
\]

\[
=\int_{-\infty}^{+\infty}\exp\left[-2\pi j\omega_{2}z_{2}\right]\left\{ \int_{-\infty}^{+\infty}\exp\left[-2\pi j\omega_{1}z_{1}\right]\left[1+\left(q-1\right)\left(\beta z_{1}^{2}+\beta z_{2}^{2}\right)\right]^{^{\frac{1}{1-q}}}dz_{1}\right\} dz_{2}
\]

\[
=\int_{-\infty}^{+\infty}\exp\left[-2\pi j\omega_{2}z_{2}\right]\left\{ \int_{-\infty}^{+\infty}\exp\left[-2\pi j\omega_{1}z_{1}\right]\left[1+\left(q-1\right)\beta z_{2}^{2}+\left(q-1\right)\beta z_{1}^{2}\right]^{^{\frac{1}{1-q}}}dz_{1}\right\} dz_{2}
\]

\[
=\int_{-\infty}^{+\infty}\exp\left[-2\pi j\omega_{2}z_{2}\right]\times
\]

\[
\left\{ \int_{-\infty}^{+\infty}\exp\left[-2\pi j\omega_{1}z_{1}\right]\left[1+\left(q-1\right)\beta z_{2}^{2}\right]^{\frac{1}{1-q}}\left[1+\left(\frac{q-1}{1+\left(q-1\right)\beta z_{2}^{2}}\right)\beta z_{1}^{2}\right]^{^{\frac{1}{1-q}}}dz_{1}\right\} dz_{2}
\]

\[
=\int_{-\infty}^{+\infty}\exp\left[-2\pi j\omega_{2}z_{2}\right]\left[1+\left(q-1\right)\beta z_{2}^{2}\right]^{\frac{1}{1-q}}\times
\]

\[
\left\{ \int_{-\infty}^{+\infty}\exp\left[-2\pi j\omega_{1}z_{1}\right]\left[1+\left(q-1\right)\left(\frac{\beta}{1+\left(q-1\right)\beta z_{2}^{2}}\right)z_{1}^{2}\right]^{^{\frac{1}{1-q}}}dz_{1}\right\} dz_{2}
\]

\[
=\int_{-\infty}^{+\infty}\exp\left[-2\pi j\omega_{2}z_{2}\right]\left[1+\left(q-1\right)\beta z_{2}^{2}\right]^{\frac{1}{1-q}}\times
\]

\[
\left\{ \int_{-\infty}^{+\infty}\exp\left[-2\pi j\omega_{1}z_{1}\right]\left[1+\left(q-1\right)\zeta z_{1}^{2}\right]^{^{\frac{1}{1-q}}}dz_{1}\right\} dz_{2},
\]

where:

\[
\zeta=\left(\frac{\beta}{1+\left(q-1\right)\beta z_{2}^{2}}\right).
\]

Let

\[
F_{1}\left(\omega_{1},\zeta\left(z_{2},q\right),q\right)=\int_{-\infty}^{+\infty}\exp\left[-2\pi j\omega_{1}z_{1}\right]\left[1+\left(q-1\right)\zeta z_{1}^{2}\right]^{^{\frac{1}{1-q}}}dz_{1}.
\]

Therefore, we can return to expression (\ref{eq:2D-QGaussian-FT01})
and write:

\begin{equation}
\mathcal{F}\left(G_{2,q}\left(\mathbf{z},\Sigma,\beta\right);\omega_{1},\omega_{2}\right)=C_{2,q}\left(\Sigma,\beta\right)\sqrt{|\Sigma|}\times
\end{equation}

\begin{equation}
\int_{-\infty}^{+\infty}\exp\left[-2\pi j\omega_{2}z_{2}\right]\left[1+\left(q-1\right)\beta z_{2}^{2}\right]^{\frac{1}{1-q}}F_{1}\left(\omega_{1},\zeta\left(z_{2},q\right),q\right)dz_{2}\label{eq:2D-QGaussian-FT002}
\end{equation}

By using the property that the Fourier transform of the product of
two functions can be computed by the convolution in the Fourier transform
domain, we can re-write expression (\ref{eq:2D-QGaussian-FT002})
as:

\begin{equation}
\mathcal{F}\left(G_{2,q}\left(\mathbf{z},\beta\right);\omega_{1},\omega_{2}\right)=
\end{equation}

\begin{equation}
=C_{2,q}\left(\Sigma,\beta\right)\sqrt{|\Sigma|}\left(\int_{-\infty}^{+\infty}\exp\left[-2\pi j\omega_{2}z_{2}\right]\left[1+\left(q-1\right)\beta z_{2}^{2}\right]^{\frac{1}{1-q}}dz_{2}\right)\otimes
\end{equation}

\begin{equation}
\left(\int_{-\infty}^{+\infty}\exp\left[-2\pi j\omega_{2}z_{2}\right]F_{1}\left(\omega_{1},a\left(z_{2},q\right),q\right)dz_{2}\right).\label{eq:2D-QGaussian-FT003}
\end{equation}

Unfortunately, this expression can not be analytically resolved like
in the one-dimensional case. Therefore, we apply numerical methods
to approximate it, as done next.

\section{Numerical Computation of FT\label{sec:Numerical-Computation-FT2D}}

We consider the isotropic setup ($\sigma_{1}=\sigma_{2}=\sigma$)
and approximate expression (\ref{eq:2D-QGaussian-FT00}) by the double
summation:

\begin{equation}
\mathcal{F}\left(G_{2,q}\left(\mathbf{x},\beta\right);\omega_{1},\omega_{2}\right)\approx\sum_{m=-\infty}^{+\infty}\sum_{n=-\infty}^{+\infty}\exp\left[-2\pi j\left(\omega_{1}x_{m}+\omega_{2}y_{n}\right)\right]G_{2,q}\left(x_{m},y_{n};\beta\right)\Delta x\Delta y,\label{eq:FT-2D-q-gaussian-discrete00}
\end{equation}
where $\Delta x$ and $\Delta y$ must be chosen in advance. From
expressions (\ref{eq:final-2D-gaussian00-1})-(\ref{eq:final-2D-gaussian01-1})
it is straightforward to show that $G_{2,q}\left(\mathbf{x},\beta\right)\rightarrow0$
if $||\mathbf{x}||\rightarrow+\infty$. Consequently, we can limit
the support of $G_{2,q}$ in a rectangular region $D\subset\mathbb{R}^{2}$and
replace the infinite double summations by finite ones whenever $\Delta x,\Delta y>0$.
The numerical error in this process is controlled by the sizes of $D$
and discretization parameters $\Delta x,\Delta y$. Once $G_{2,q}\left(\mathbf{x},\beta\right)$
is a symmetric function around the origin $\left(0,0\right)$we can
place the center of the region $D$ in the origin $\left(0,0\right)$.
Besides, in the isotropic case we can also postulate $\Delta x=\Delta y=T$
and use a square region $D$. Hence, the numerical setup to compute
expression (\ref{eq:FT-2D-q-gaussian-discrete00}) depends on the
definition of $T$ and the size of $D$ . Also, in this case we have
$x_{m}=mT$ and $y_{n}=nT$ in expression (\ref{eq:FT-2D-q-gaussian-discrete00})
and we can find a square $D$ in the space domain such that $G_{2,q}\left(\mathbf{x},\beta\right)<\delta$
if $\mathbf{x}\notin D$, for a given $\delta>0$. By putting all
these considerations together, we can re-write expression (\ref{eq:FT-2D-q-gaussian-discrete00})
as:

\begin{equation}
\mathcal{\overline{F}}\left(G_{2,q}\left(\mathbf{x},\beta\right);\omega_{1},\omega_{2}\right)\approx\sum_{m=-M}^{M}\sum_{n=-M}^{M}\exp\left[-2\pi j\left(\omega_{1}x_{m}+\omega_{2}y_{n}\right)\right]G_{2,q}\left(x_{m},y_{n};\beta\right)T^{2},\label{eq:FT-2D-q-gaussian-discrete00-1}
\end{equation}
where $-T^{-1}<\omega_{1},\omega_{2}<T^{-1}$, $x_{m}=mT$, $y_{n}=nT$, with $-M\leq m,n\leq M$. It is important
to be aware to the fact that expression (\ref{eq:FT-2D-q-gaussian-discrete00-1})
has a period $T^{-1}$ in both frequency directions $\omega_{1}$
and $\omega_{2}$. In fact, if we compute:

\[
\mathcal{\overline{F}}\left(G_{2,q}\left(\mathbf{x},\beta\right);\omega_{1}+T^{-1},\omega_{2}+T^{-1}\right)
\]

\[
=\sum_{m=-M}^{M}\sum_{n=-M}^{M}\exp\left[-2\pi j\left(\omega_{1}x_{m}+\omega_{2}y_{n}\right)-2\pi j\left(x_{m}+y_{n}\right)T^{-1}\right]G_{2,q}\left(x_{m},y_{n};\beta\right)T^{2}
\]

\[
=\sum_{m=-M}^{M}\sum_{n=-M}^{M}\exp\left[-2\pi j\left(\omega_{1}x_{m}+\omega_{2}y_{n}\right)\right]\exp\left[-2\pi j\left(m+n\right)\right]G_{2,q}\left(x_{m},y_{n};\beta\right)T^{2}
\]

\[
=\sum_{m=-M}^{M}\sum_{n=-M}^{M}\exp\left[-2\pi j\left(\omega_{1}x_{m}+\omega_{2}y_{n}\right)\right]G_{2,q}\left(x_{m},y_{n};\beta\right)T^{2},
\]
once $m,n$ are integer numbers. That is why we must compute expression
(\ref{eq:FT-2D-q-gaussian-discrete00-1}) only inside the square $-T^{-1}\leq\omega_{1},\omega_{2}\leq T^{-1}$
in the frequency domain.

\section{Computational Experiments\label{sec:Computational-Experiments}}

Before proceeding, we shall notice that expressions (\ref{eq:FT-of-q-gaussian-q-maior-1})-(\ref{eq:FT-of-q-gaussian-q-menor-1-y-0})
need some considerations before their computations. Expression (\ref{eq:FT-of-q-gaussian-q-maior-1})
involves the modified Bessel functions, given by equation (\ref{eq:degenerate-hypergeometric}),
which is not defined by $\mu\in\mathbb{Z}_{-}$. Also, the Gamma function,
defined by expression (\ref{eq:gamma-function-definition-Euler})
is not valid for $z\in\mathbb{Z}_{-}$, which imposes constraints
for the Bessel functions also (see expression (\ref{eq:Bessel-functions})).
We must notice that the Gamma function occurs also in the normalization
factor $C_{1,q}$ given in expressions (\ref{eq:normalization-constant-q-gaussian00})-(\ref{eq:normalization-constant-q-gaussian01}).
Therefore, if we put all the constraints together, we conclude that
these expression can be computed only if:

a) Case $1<q<3$

\[
\left\{ \frac{1}{q-1}+\frac{1}{2}\right\} \cap\mathbb{Z}=\emptyset,and\quad\left\{ \frac{1}{q-1}\right\} \cap\mathbb{Z}_{-}=\emptyset.
\]

b) Case $q<1$

\[
\left\{ \frac{1}{1-q}+1\right\} \cap\mathbb{Z}_{-}=\left\{ \frac{1}{1-q}+\frac{1}{2}\right\} \cap\mathbb{Z}_{-}=\left\{ \frac{1}{1-q}+\frac{3}{2}\right\} \cap\mathbb{Z}_{-}=\emptyset.
\]

The center $x^{\ast}$ and radius $\Delta_{\psi}$ \ of a function
$\psi=\psi\left(x\right)$ are defined to be:
\begin{equation}
x^{\ast}:=\frac{1}{\left\Vert \psi\right\Vert _{2}^{2}}\overset{\infty}{\underset{-\infty}{\int}}x\,\left|\psi\left(x\right)\right|^{2}dx,\label{win 00}
\end{equation}
\begin{equation}
\Delta_{\psi}\equiv\frac{1}{\left\Vert \psi\right\Vert _{2}}\left\Vert \left(x-x^{\ast}\right)\psi\right\Vert _{2}=\frac{1}{\left\Vert \psi\right\Vert _{2}}\left\{ \overset{\infty}{\underset{-\infty}{\int}}\left(x-x^{\ast}\right)^{2}\,\left|\psi\left(x\right)\right|^{2}dx\right\} ^{1/2}.\label{win 01}
\end{equation}

\begin{figure}[!htb]
\begin{centering}
\begin{minipage}[b]{7cm}%
\begin{center}
\includegraphics[width=1\linewidth]{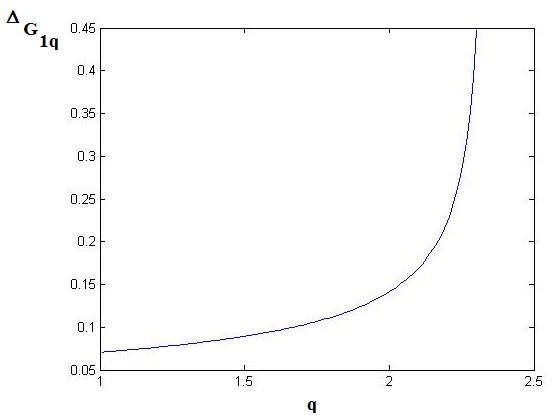} (a)
\par\end{center}%
\end{minipage}%
\begin{minipage}[b]{7cm}%
\begin{center}
\includegraphics[width=1\linewidth]{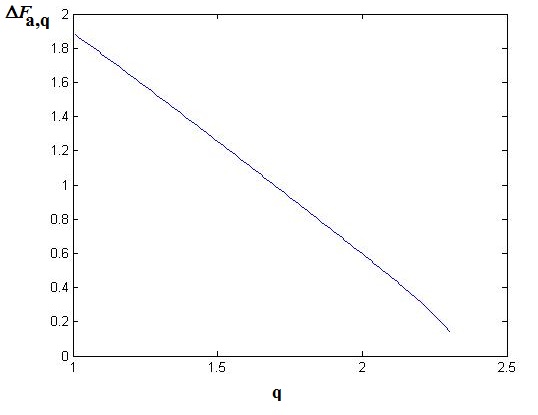}
(b)
\par\end{center}%
\end{minipage}
\par\end{centering}
\caption{(a) Size of space window for $q$-Gaussian with parameters $\beta=1/2$,$\sigma=0.1$.
(b) Behavior of expression (\ref{eq:tendency-frequency-window}) for
$q$-Gaussian with $\beta=1/2$,$\sigma=0.1$.}
\label{fig:window-size}
\end{figure}

If $x^{\ast}<\infty$ and $\Delta_{\psi}<\infty$, we say that the
signal $\psi$ is \textit{localized} about the point $t^{\ast}$ with
the space window $\left[x^{\ast}-\Delta_{\psi},x^{\ast}+\Delta_{\psi}\right].$
The space window corresponding to the one-dimensional $q$-Gaussian,
given by expression (\ref{eq:Q-Gaussian-d-1-1}), can be computed
by noticing that $x^{\ast}=0$ due to the fact that $G_{1,q}\left(x,\sigma,\beta\right)=G_{1,q}\left(-x,\sigma,\beta\right)$.
Besides, we can use a methodology that is analogous to the one applied
in Appendix \ref{sec:Appendix-C-One-Dimensional} to show that, for $q>1$ we get:

\begin{equation}
\left\Vert G_{1,q}\left(x,\sigma,\beta\right)\right\Vert _{2}\equiv\left\{ \overset{\infty}{\underset{-\infty}{\int}}\left|G_{1,q}\left(x,\sigma,\beta\right)\right|^{2}dx\right\} ^{1/2}=\frac{C_{1q}\left(\sigma,\beta\right)}{\left[\left(q-1\right)\frac{\beta}{\sigma^{2}}\right]^{1/4}}\sqrt{B\left(\frac{1}{2},\frac{2}{q-1}-\frac{1}{2}\right)},\label{eq:window00}
\end{equation}
with $1<q<3$,

\begin{equation}
\left\Vert xG_{1,q}\left(x,\sigma,\beta\right)\right\Vert _{2}=\frac{C_{1q}\left(\sigma,\beta\right)}{\left[\left(q-1\right)\frac{\beta}{\sigma^{2}}\right]^{3/4}}\sqrt{B\left(\frac{3}{2},\frac{2}{q-1}-\frac{3}{2}\right)},\quad1<q<\frac{7}{3},\label{eq:window01}
\end{equation}

\begin{equation}
\Delta_{G_{1,q}}^{\sigma,\beta}=\left(\frac{B\left(\frac{3}{2},\frac{2}{q-1}-\frac{3}{2}\right)}{\left[\left(\left(q-1\right)\frac{\beta}{\sigma^{2}}\right)\right]B\left(\frac{1}{2},\frac{2}{q-1}-\frac{1}{2}\right)}\right)^{1/2},\quad1<q<\frac{7}{3},\label{eq:window02}
\end{equation}
where $C_{1q}\left(\sigma,\beta\right)$ is given by expression (\ref{eq:normalization-constant-q-gaussian00}),
and $B$ is the Beta function given by equation (\ref{eq:def-function-beta}).
In the section \ref{sub:C.3-One-Dimensional-Q-Gaussian-Expression}
we show that for $\beta=1/2$ the $q$-Gaussian generates the traditional
Gaussian in the limit $q\rightarrow1$. Therefore, to allow further
comparisons, we set $\beta=1/2$, and arbitrarily choose $\sigma=0.1$
expression (\ref{eq:Q-Gaussian-d-1-1}). The Figure \ref{fig:window-size}.(a)
shows the behaviour of expression (\ref{eq:window02}) in the corresponding
$q$ range.

\begin{figure}[!htb]
\begin{centering}
\begin{minipage}[b]{7cm}%
\begin{center}
\includegraphics[width=1\linewidth]{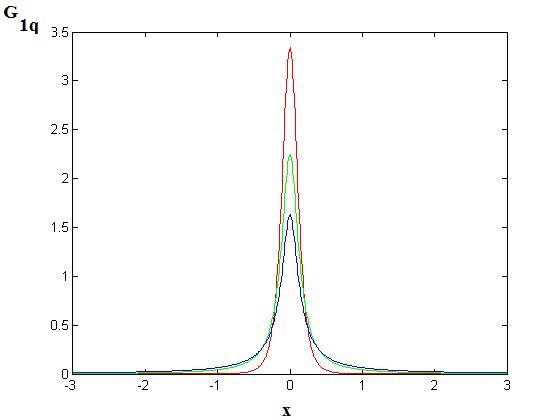}
(a)
\par\end{center}%
\end{minipage}%
\begin{minipage}[b]{7cm}%
\begin{center}
\includegraphics[width=1\linewidth]{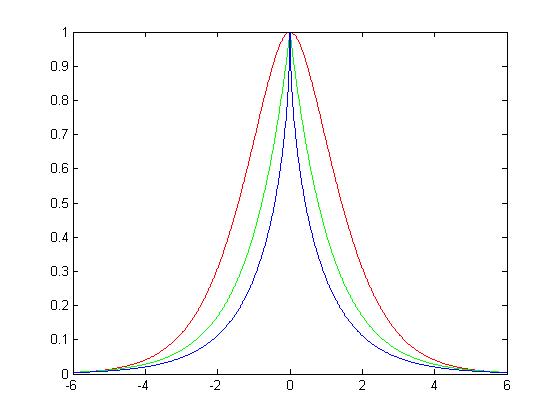}
(b)
\par\end{center}%
\end{minipage}
\par\end{centering}
\caption{(a) Plot for $q$-Gaussian in the space domain with parameters $\beta=1/2$,$\sigma=0.1$
and $q=1.41$ (red), $q=2.0$ (green) and $q=2.3$ (blue). (b) Absolute
value of the Fourier transform of $q$-Gaussian (equation (\ref{eq:abs-FT-qGaussian-q-maior-1}))
with parameters $\beta=1/2$,$\sigma=0.1$ and $q=1.41$ (red), $q=2.0$
(green) and $q=2.3$ (blue).}

\label{fig:q-gaussian-space-frequency-domains}
\end{figure}

We notice that the size of the space window of $G_{1,q}\left(x,0.1,0.5\right)$
about $x^{\ast}=0$ is a monotone increasing function respect to $q$.
Consequently, we expect the behavior shown in Figure \ref{fig:q-gaussian-space-frequency-domains}.(a).
Also, due to the Heisenberg inequality:

\begin{equation}
4\pi\left\Vert xG_{1,q}\left(x,\sigma,\beta\right)\right\Vert _{2}\cdot\left\Vert y\mathcal{F}\left(G_{1,q}\left(x,\sigma,\beta\right);y\right)\right\Vert _{2}\geq\left\Vert G_{1,q}\left(x,\sigma,\beta\right)\right\Vert _{2}^{2},\label{eq:Heisenberg inequality}
\end{equation}
we get that:

\[
\left\Vert y\mathcal{F}\left(G_{1,q}\left(x,\sigma,\beta\right);y\right)\right\Vert _{2}\geq\Delta\mathcal{F}_{\sigma,\beta,q},
\]
where:

\begin{equation}
\Delta\mathcal{F}_{1,q}^{\sigma,\beta}=\frac{C_{1q}\left[\left(q-1\right)\frac{\beta}{\sigma^{2}}\right]^{1/4}}{4\pi}\frac{B\left(\frac{1}{2},\frac{2}{q-1}-\frac{1}{2}\right)}{\sqrt{B\left(\frac{3}{2},\frac{2}{q-1}-\frac{3}{2}\right)}},\quad1<q<\frac{7}{3},\label{eq:tendency-frequency-window}
\end{equation}
and $C_{1q}$and $B$ follows like in equations (\ref{eq:window00})-(\ref{eq:window02}).
The Figure \ref{fig:window-size}.(b) indicates that the window size
in the frequency domain is a monotone decreasing function respect
to $q$. We can check this fact through Figure \ref{fig:q-gaussian-space-frequency-domains}.(b)
which pictures the profile of the absolute value of the Fourier transform
of a $q$-Gaussian for three values of the entropic index $q$. We can
notice that when increasing $q$ the $q$-Gaussian becomes more localized
about $y=0$, which agrees with the decreasing behavior pictured in
Figure \ref{fig:window-size}.(b) for expression (\ref{eq:tendency-frequency-window}).

An analogous analysis can be performed for $q<1$. In this case, we
get also $x^{\ast}=0$ from equation (\ref{win 00}) and we can use
a the same methodology applied in the Appendix \ref{sec:Appendix-C-One-Dimensional}
to get that:

\begin{equation}
\left\Vert G_{1,q}\left(x,\sigma,\beta\right)\right\Vert _{2}\equiv\left\{ \overset{\infty}{\underset{-\infty}{\int}}\left|G_{1,q}\left(x,\sigma,\beta\right)\right|^{2}dx\right\} ^{1/2}=\frac{C_{1q}}{\left[\left(1-q\right)\frac{\beta}{\sigma^{2}}\right]^{1/4}}\sqrt{B\left(\frac{1}{2},\frac{2}{1-q}+1\right)},\quad q<1,\label{eq:window00-1}
\end{equation}

\begin{equation}
\left\Vert xG_{1,q}\left(x,\sigma,\beta\right)\right\Vert _{2}=\frac{C_{1q}}{\left[\left(1-q\right)\frac{\beta}{\sigma^{2}}\right]^{3/4}}\sqrt{B\left(\frac{3}{2},\frac{2}{1-q}+1\right)},\quad q<1,\label{eq:window01-1}
\end{equation}

\begin{equation}
\Delta_{G_{1,q}}^{\sigma,\beta}=\left(\frac{B\left(\frac{3}{2},\frac{2}{1-q}+1\right)}{\left[\left(\left(1-q\right)\frac{\beta}{\sigma^{2}}\right)\right]B\left(\frac{1}{2},\frac{2}{1-q}+1\right)}\right)^{1/2},\quad q<1,\label{eq:window02-1}
\end{equation}
where $|\mathbf{x}|\leq\left(\left(1-q\right)\frac{\beta}{\sigma^{2}}\right)^{-1/2}$,
$C_{1q}$ is given by expression (\ref{eq:normalization-constant-q-gaussian01}),
and $B$ is the Beta function, computed by equation (\ref{eq:def-function-beta}).
By using Heisenberg inequality, given by expression (\ref{eq:Heisenberg inequality}),
we can obtain:

\[
\left\Vert y\mathcal{F}\left(G_{1,q}\left(x,\sigma,\beta\right);y\right)\right\Vert _{2}\geq\Delta\mathcal{F}_{\sigma,\beta,q},
\]
with:

\begin{equation}
\Delta\mathcal{F}_{1,q}^{\sigma,\beta}=\frac{C_{1q}}{4\pi}\left[\left(1-q\right)\frac{\beta}{\sigma^{2}}\right]^{1/4}\frac{B\left(\frac{1}{2},\frac{2}{1-q}+1\right)}{\sqrt{B\left(\frac{3}{2},\frac{2}{1-q}+1\right)}},\quad q<1\label{eq:Hisemberg-bound-q-menor-1}
\end{equation}

\begin{figure}[!htb]
\begin{centering}
\begin{minipage}[b]{7cm}%
\begin{center}
\includegraphics[width=1\linewidth]{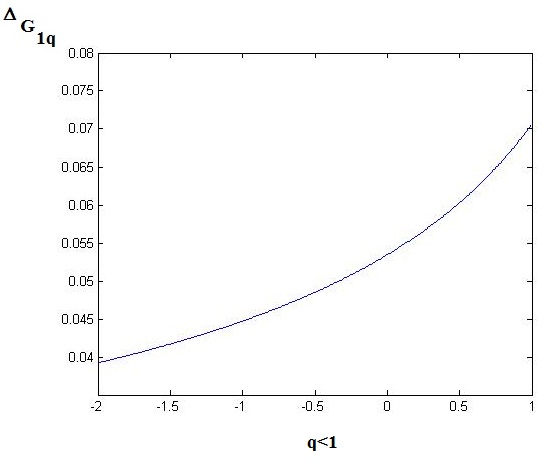}
(a)
\par\end{center}%
\end{minipage}%
\begin{minipage}[b]{7cm}%
\begin{center}
\includegraphics[width=1\linewidth]{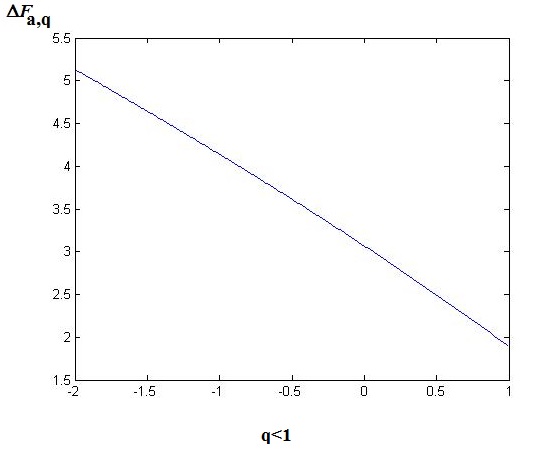}
(b)
\par\end{center}%
\end{minipage}
\par\end{centering}
\caption{(a) Size of space window for $q$-Gaussian with parameters $\beta=1/2$,$\sigma=0.1$
for $q<1$. (b) Behavior of expression (\ref{eq:Hisemberg-bound-q-menor-1})
for $q$-Gaussian with $\beta=1/2$,$\sigma=0.1$.}
\label{fig:window-size-q-menor-1}
\end{figure}

The Figure \ref{fig:window-size-q-menor-1}.(a) shows the behaviour
of expression (\ref{eq:window02-1}) in the range $-2.0<q<0.99$ and
the Figure \ref{fig:window-size-q-menor-1}.(b) pictures the behavior
of expression (\ref{eq:Hisemberg-bound-q-menor-1}) in the same range.
Likewise in the $q>1$case, we observe that $\Delta_{G_{1,q}}^{0.5,0.1}$and
$\Delta\mathcal{F}_{1,q}^{0.5,0.1}$ are monotone increasing and monotone
decreasing functions, respectively, in the considered $q$ range.

\begin{figure}[!htb]
\begin{centering}
\begin{minipage}[b]{7cm}%
\begin{center}
\includegraphics[width=1\linewidth]{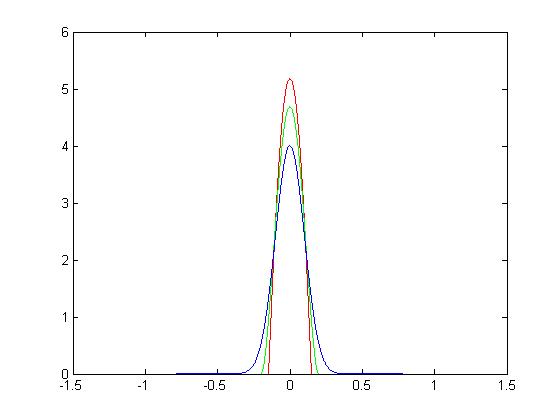}
(a)
\par\end{center}%
\end{minipage}%
\begin{minipage}[b]{7cm}%
\begin{center}
\includegraphics[width=1\linewidth]{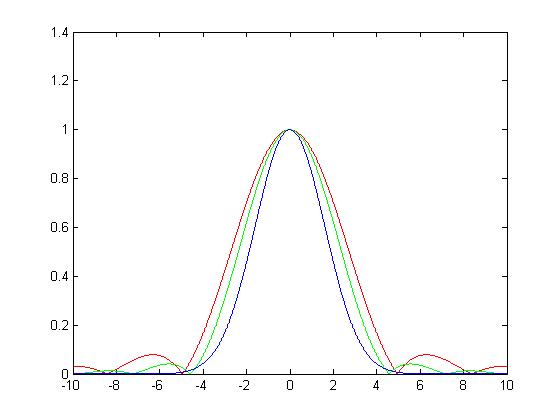}
(b)
\par\end{center}%
\end{minipage}
\par\end{centering}
\caption{(a) Plot for $q$-Gaussian in the space domain with parameters $\beta=1/2$,$\sigma=0.1$
and $q=0.1$ (red), $q=0.5$ (green) and $q=0.99$ (blue). (b) FT
of $q$-Gaussian (denoted by$\mathcal{F}\left(G_{1,q}\left(x,a\right);y\right)$)
with parameters $\beta=1/2$,$\sigma=0.1$ and $q=0.1$ (red), $q=0.5$
(green) and $q=0.99$ (blue).}
\label{fig:q-gaussian-space-frequency-domains-q-menor-1}
\end{figure}

We can check these facts through Figure \ref{fig:q-gaussian-space-frequency-domains-q-menor-1}
which pictures the $q$-Gaussian and its Fourier transform for $q=0.1,\,0.5,\,0.99$.
We can notice by Figure \ref{fig:q-gaussian-space-frequency-domains-q-menor-1}.(a)
that when increasing $q$ the $q$-Gaussian becomes less localized about
$y=0$, which agrees with the increasing behavior of the window size
pictured in Figure \ref{fig:window-size-q-menor-1}.(a) . On the other
hand, the tendency for the Fourier transform localization given by
expression (\ref{eq:Hisemberg-bound-q-menor-1}), and represented
in the Figure \ref{fig:window-size-q-menor-1}.(b) is confirmed by
the Figure \ref{fig:q-gaussian-space-frequency-domains-q-menor-1}.(b).

The Figures \ref{fig:q-gaussian-space-frequency-domains}.(b) and
\ref{fig:q-gaussian-space-frequency-domains-q-menor-1}.(b) shows
that: $\mathcal{F}\left(G_{1,q}\left(x,\sigma,\beta\right);0\right)=1$
for the considered $q$ values. In fact, from expressions (\ref{eq:abs-FT-qGaussian-q-maior-1})-(\ref{eq:FT-of-q-gaussian-q-menor-1-y-0})
we can prove this property for any $q\in\mathbb{R}-\left\{ 1\right\} $and
make comparisons with the (normalized) Gaussian given by expression:

\begin{equation}
G\left(x,\sigma\right)=\frac{1}{\sigma\sqrt{2\pi}}\exp\left(-\frac{x^{2}}{2\sigma^{2}}\right),\label{eq:traditional-gaussian-01}
\end{equation}
whose Fourier transform is the function:

\begin{equation}
\mathcal{F}\left(G\left(x,\sigma\right);y\right)=\exp\left(-\sigma y^{2}\right).\label{eq:traditional-FT-gaussian}
\end{equation}

So, it is clear that $\mathcal{F}\left(G\left(x,\sigma\right);0\right)=1$
also. Besides, Figures \ref{fig:q-gaussian-space-frequency-domains}
and \ref{fig:q-gaussian-space-frequency-domains-q-menor-1} show that,
if we fix the parameters $\beta$ and $\sigma$ in expression (\ref{eq:Q-Gaussian-d-1-1}),
we can change the localization and the profile of the $q$-Gaussian by
changing the $q$ value. Therefore, in terms of low-pass filtering
properties, the main point is how close $G_{1,q}\left(x,\sigma,\beta\right)$
(and $\mathcal{F}\left(G_{1,q}\left(x,\sigma,\beta\right);y\right)$)
is from $G\left(x,\sigma\right)$ (and $\mathcal{F}\left(G\left(x,\sigma\right);y\right)$)
when changing the $q$ value in expression (\ref{eq:traditional-FT-gaussian})?
We must perform further developments in order to answer this question.

\begin{figure}[!htb]
\begin{centering}
\begin{minipage}[b]{7cm}%
\begin{center}
\includegraphics[width=1\linewidth]{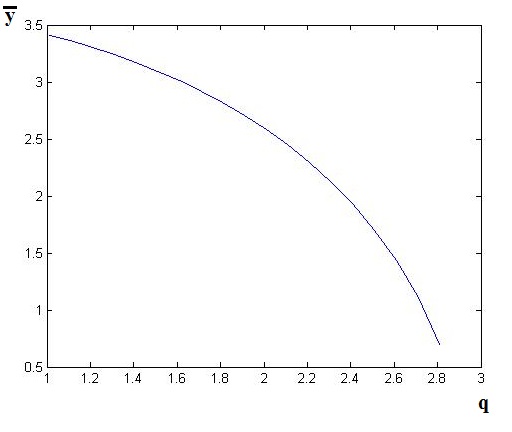}
(a)
\par\end{center}%
\end{minipage}%
\begin{minipage}[b]{7cm}%
\begin{center}
\includegraphics[width=1\linewidth]{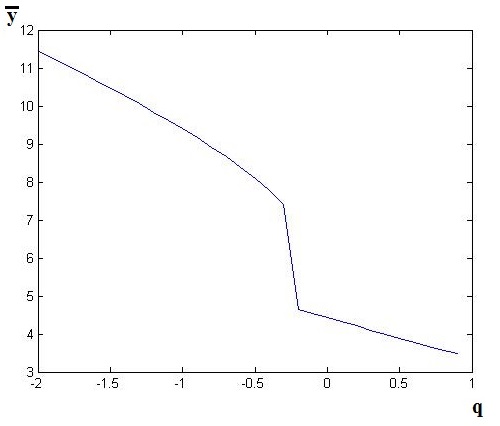}
(b)
\par\end{center}%
\end{minipage}
\par\end{centering}
\caption{(a) Cut-off frequency $\bar{y}$ such that $abs\left(\mathcal{F}\left(G_{1,q}\left(x,0.1,0.5\right);\bar{y}\right)\right)<0.1$
for $1<q<3$. (b) Cut-off frequency $\bar{y}$ such that $abs\left(\mathcal{F}\left(G_{1,q}\left(x,0.1,0.5\right);\bar{y}\right)\right)<0.1$
for $-2\leq q<1$.}
\label{fig:cutoff-frquency-q-gaussian-frequency-domain}
\end{figure}

On the other hand, Figure \ref{fig:cutoff-frquency-q-gaussian-frequency-domain}
shows the cut-off frequency $\bar{y}=\bar{y}\left(q\right)$, such
that $abs\left(\mathcal{F}\left(G_{1,q}\left(x,0.1,0.5\right);\bar{y}\right)\right)<0.1$.

We notice that $\bar{y}$ is a decreasing function which is in accordance
with the behaviour reported by Figures \ref{fig:q-gaussian-space-frequency-domains}.(a)
and \ref{fig:q-gaussian-space-frequency-domains-q-menor-1}.(a).

\begin{figure}[!htb]
\begin{centering}
\begin{minipage}[b]{8cm}%
\begin{center}
\includegraphics[width=1.5\linewidth]{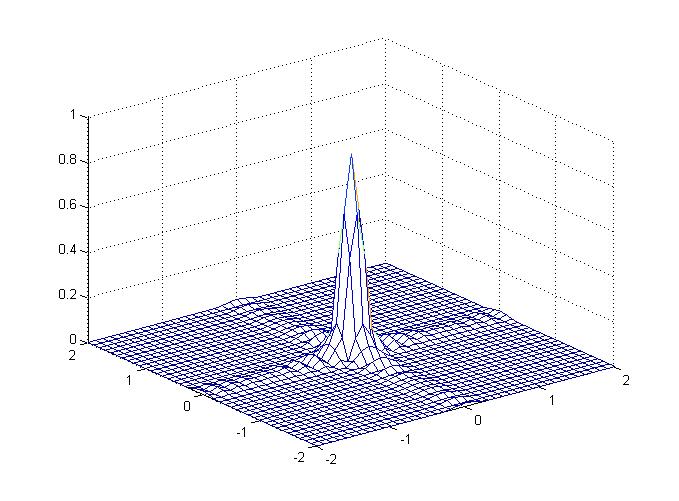}
\end{center}%
\end{minipage}
\end{centering}
\caption{Plot of $abs\left(\mathcal{F}\left(G_{2,0.5}\left(x,y;\sqrt{8},1\right);\omega_{1},\omega_{2}\right)\right)$
for $-2\leq\omega_{1},\omega_{2}<2$. }
\label{fig:abs-FT-2D-q-gaussian-frequency-domain}
\end{figure}

Now, we consider the FT of the $q$-Gaussian 2D defined by the following
parameters: $\beta=1$,$\sigma=\sqrt{8}\approx2.8284$, $q=0.5$.
The discretization parameters used in expression (\ref{eq:FT-2D-q-gaussian-discrete00-1})
are: $T=0.25$, $M=2.5$. The Figure \ref{fig:abs-FT-2D-q-gaussian-frequency-domain}
pictures the obtained surface.

\section{Conclusions and Future Works\label{sec:Conclusions-and-Future}}
In this paper we collect theoretical
elements to analyze the $q$-Gaussian kernel for feature extraction and edge detection.
We review some theoretical elements behind the $q$-Gaussian
and its Fourier transform. We analyze the $q$-Gaussian kernel in the space and Fourier
domains using the concepts of space window, cut-off frequency, and
the Heisenberg inequality. We postulate that the comparison between
the $q$-Gaussian and Gaussian kernels in the Fourier/space domains may allow
to explain the observed smoothing capabilities of the $q$-Gaussian kernel
for $q<1$.

\section*{Acknowledgement}
We would like to thank Dr. Ernesto Pinheiro Borges, from Federal
University of Bahia, for the comments and mathematical developments, which
helped us a lot to complete this work.


\appendix
\section{Appendix: Gaussian in $\mathbb{R}^{d}$\label{sec:Appendix-A-Gaussian}}

The $d$-dimensional Gaussian is defined by:

\begin{equation}
G\left(\mathbf{x},\Sigma\right)=\left[\left(2\pi\right)^{d/2}|\Sigma|^{1/2}\right]^{-1}\exp\left(-\frac{1}{2}\mathbf{x^{T}}\Sigma^{-1}\mathbf{x}\right).\label{eq:gaussian-2d}
\end{equation}
where $\Sigma$ is the covariance matrix (symmetric and positive definite)
and $|\Sigma|$ means its determinant.

If $d=1:$

\[
G\left(x,\sigma^{2}\right)=\left[\sqrt{2\pi\sigma^{2}}\right]^{-1}\exp\left(-\frac{x^{2}}{2\sigma^{2}}\right)=\frac{1}{\sigma\sqrt{2\pi}}\exp\left(-\frac{x^{2}}{2\sigma^{2}}\right)
\]

Analogously, if $d=2$ then $\mathbf{x}=\left(x,y\right)$ and expression
(\ref{eq:gaussian-2d}) renders:

\[
G\left(x,y,\Sigma\right)=\left[\left(2\pi\right)|\Sigma|^{1/2}\right]^{-1}\exp\left(-\frac{1}{2}\left(\begin{array}{cc}
x & y\end{array}\right)\Sigma^{-1}\left(\begin{array}{c}
x\\
y
\end{array}\right)\right).
\]

If the covariance matrix is diagonal:

\[
\Sigma=\left(\begin{array}{cc}
\sigma_{1}^{2} & 0\\
0 & \sigma_{2}^{2}
\end{array}\right)\quad\Longrightarrow\quad\Sigma^{-1}=\left(\begin{array}{cc}
\sigma_{1}^{-2} & 0\\
0 & \sigma_{2}^{-2}
\end{array}\right).
\]

So:

\[
G\left(x,y,\Sigma\right)=\frac{1}{2\pi\sigma_{1}\sigma_{2}}\exp\left(-\frac{1}{2}\left(\begin{array}{cc}
x & y\end{array}\right)\left(\begin{array}{cc}
\sigma_{1}^{-2} & 0\\
0 & \sigma_{2}^{-2}
\end{array}\right)\left(\begin{array}{c}
x\\
y
\end{array}\right)\right),
\]
and:

\[
G\left(x,y,\Sigma\right)=\frac{1}{2\pi\sigma_{1}\sigma_{2}}\exp\left(-\frac{1}{2}\left(\frac{x^{2}}{\sigma_{1}^{2}}+\frac{y^{2}}{\sigma_{2}^{2}}\right)\right).
\]

\section{Appendix: $q$-Gaussian in $\mathbb{R}^{d}$\label{sec:Appendix-B-Q-Gaussian}}

The $d$-dimensional $q$-Gaussian is defined by:

\begin{equation}
G_{d,q}\left(\mathbf{x},\Sigma,\beta\right)=C_{d,q}\left(\Sigma,\beta\right)\exp_{q}\left(-\beta\mathbf{x^{T}}\Sigma^{-1}\mathbf{x}\right),\label{eq:Q-Gaussian-d-2}
\end{equation}
where $C_{d,q}\left(\Sigma,\beta\right)$ is a normalization factor.

Once the covariance matrix $\Sigma$ is symmetric and positive definite,
there exists an orthogonal matrix $U$ and a diagonal matrix $D$
such that:

\[
\Sigma=U^{T}DU.
\]

Let

\begin{equation}
\mathbf{y}=U\mathbf{x},\label{eq:defining-y}
\end{equation}
.

Then $\mathbf{x}=U^{T}\mathbf{y}$ and, once $abs|U|=1:$

\[
C_{d,q}\left(\Sigma,\beta\right)=\left(\int_{\mathbb{R}^{d}}\exp_{q}\left(-\beta\mathbf{y^{T}}U\Sigma^{-1}U^{T}\mathbf{y}\right)d\mathbf{y}\right)^{-1}.
\]

But:

\begin{equation}
\Sigma^{-1}=\left(U^{T}DU\right)^{-1}=U^{T}D^{-1}U\quad\Longrightarrow\quad D^{-1}=U\Sigma^{-1}U^{T}.\label{eq:inv-d-sigma}
\end{equation}

Therefore:

\[
C_{d,q}\left(\Sigma,\beta\right)=\left(\int_{\mathbb{R}^{d}}\exp_{q}\left(-\beta\mathbf{y^{T}}D^{-1}\mathbf{y}\right)d\mathbf{y}\right)^{-1}.
\]

If:

\[
D^{-1}=\left(\begin{array}{cccc}
\sigma_{1}^{-2}\\
 & \sigma_{2}^{-2}\\
 &  & ...\\
 &  &  & \sigma_{d}^{-2}
\end{array}\right),
\]

then: $D^{-1}=\sqrt{D^{-1}}\sqrt{D^{-1}}$, where:

\[
\sqrt{D^{-1}}=\left(\begin{array}{cccc}
\sigma_{1}^{-1}\\
 & \sigma_{2}^{-1}\\
 &  & ...\\
 &  &  & \sigma_{d}^{-1}
\end{array}\right).
\]

If

\begin{equation}
\mathbf{z}=\sqrt{D^{-1}}\mathbf{y},\label{eq:defining-z}
\end{equation}

then

\[
d\mathbf{z}=det\left(\sqrt{D^{-1}}\right)d\mathbf{y},
\]

and;

\[
det\left(\sqrt{D^{-1}}\right)=det\left(D^{-1/2}\right)=\left(\frac{1}{\sqrt{det\left(D\right)}}\right)=\frac{1}{\sqrt{det\left(U\Sigma U^{T}\right)}}=\frac{1}{\sqrt{det\left(\Sigma\right)}}\equiv\frac{1}{\sqrt{|\Sigma|}}.
\]

Therefore:

\[
d\mathbf{y}=\sqrt{|\Sigma|}d\mathbf{z}.
\]

Consequently:

\begin{equation}
C_{d,q}\left(\Sigma,\beta\right)=\left(\int_{\mathbb{R}^{d}}\exp_{q}\left(-\beta\mathbf{y^{T}}D^{-1}\mathbf{y}\right)d\mathbf{y}\right)^{-1}=\left(\int_{\mathbb{R}^{d}}\exp_{q}\left(-\beta|\mathbf{z|}^{2}\right)\sqrt{|\Sigma|}d\mathbf{z}\right)^{-1}
\end{equation}

\begin{equation}
=\frac{1}{\sqrt{|\Sigma|}\int_{\mathbb{R}^{d}}\exp_{q}\left(-\beta|\mathbf{z|}^{2}\right)d\mathbf{z}}.\label{eq:normalization-factor}
\end{equation}

\section{Appendix: One-Dimensional $q$-Gaussian \label{sec:Appendix-C-One-Dimensional}}

If $d=1$ in expression (\ref{eq:normalization-factor}) we get:

\begin{equation}
C_{1,q}\left(\sigma,\beta\right)=\left(\int_{\mathbb{R}}\exp_{q}\left(-\beta y\sigma^{-2}y\right)dy\right)^{-1}=\left(\int_{\mathbb{R}}\exp_{q}\left(-\frac{\beta}{\sigma^{2}}y^{2}\right)dy\right)^{-1}=\frac{1}{\int_{\mathbb{R}}\exp_{q}\left(-\frac{\beta}{\sigma^{2}}y^{2}\right)dy},\label{eq:normalization-factor-1}
\end{equation}
where:

\[
\int_{\mathbb{R}}\exp_{q}\left(-\frac{\beta}{\sigma^{2}}y^{2}\right)dy=\int_{\mathbb{R}}\left[1+\left(1-q\right)\left(-\frac{\beta}{\sigma^{2}}y^{2}\right)\right]^{\frac{1}{1-q}}dy
\]

\begin{equation}
=\int_{\mathbb{R}}\left[1+\left(q-1\right)\left(\frac{\beta}{\sigma^{2}}y^{2}\right)\right]^{\frac{1}{1-q}}dy.\label{eq:integral-normalization-factor-1-d}
\end{equation}

\subsection{Case $d=1$ and $q>1$}

If

\[
\tilde{y}=\left[\left(q-1\right)a\right]^{1/2}y,
\]
then:

\begin{equation}
\int_{\mathbb{R}}\exp_{q}\left(-\frac{\beta}{\sigma^{2}}y^{2}\right)dy=\int_{\mathbb{R}}\left[1+\left(q-1\right)\left(ay^{2}\right)\right]^{\frac{1}{1-q}}dy=\frac{2}{\left[\left(q-1\right)a\right]^{1/2}}\int_{0}^{+\infty}\left[1+\tilde{y}^{2}\right]^{\frac{1}{1-q}}d\widetilde{y}.\label{eq:normalizatin-for-d-1}
\end{equation}

From equations 3.251-2 and 8.384 of reference \cite{gradshteyn1981},
we get:

\begin{equation}
\int_{0}^{+\infty}x^{\mu-1}\left[1+x^{2}\right]^{\nu-1}dx=\frac{1}{2}B\left(\frac{\mu}{2},1-\nu-\frac{\mu}{2}\right),\quad if\quad Re\left(\mu\right)>0,\quad Re\left(\nu+\frac{1}{2}\mu\right)<1,\label{eq:formula-para-norm-d-1}
\end{equation}
where $B$ is the Beta function (see equation (\ref{eq:def-function-beta})).
Therefore, we can cast expression (\ref{eq:normalizatin-for-d-1})
in the above form by setting:

\[
\mu=1,\quad\nu-1=\frac{1}{1-q},
\]
which implies that:

\[
\nu=\frac{1}{1-q}+1=\frac{2-q}{1-q}.
\]

By inserting these values in the conditions for expression (\ref{eq:formula-para-norm-d-1})
we get:

\[
\frac{1}{1-q}+1+\frac{1}{2}<1,
\]
and, consequently:

\[
1<q<3.
\]

So, we can insert the above results in expression (\ref{eq:formula-para-norm-d-1})
to obtain:

\[
\int_{0}^{+\infty}\left[1+\tilde{y}^{2}\right]^{\frac{1}{1-q}}d\widetilde{y}=\int_{0}^{+\infty}\left[1+x^{2}\right]^{\frac{1}{1-q}}dx=\frac{1}{2}B\left(\frac{1}{2},\frac{1}{q-1}-\frac{1}{2}\right),
\]
where, by using the Beta function definition given by expression (\ref{eq:def-function-beta}),
it can be written:

\[
B\left(\frac{1}{2},\frac{1}{q-1}-\frac{1}{2}\right)=\frac{\Gamma\left(\frac{1}{2}\right)\Gamma\left(\frac{1}{q-1}-\frac{1}{2}\right)}{\Gamma\left(\frac{1}{q-1}\right)}.
\]

Through this result and the fact $\Sigma=\sigma^{2}$ for $d=1$,
we can compute expression (\ref{eq:normalization-factor-1}) as:

\[
C_{1,q}\left(\sigma,\beta\right)=\frac{1}{\int_{\mathbb{R}}\exp_{q}\left(-\frac{\beta}{\sigma^{2}}y^{2}\right)dy}
\]

\begin{equation}
=\frac{1}{\frac{2}{\left[\left(q-1\right)a\right]^{1/2}}\frac{1}{2}\frac{\Gamma\left(\frac{1}{2}\right)\Gamma\left(\frac{1}{q-1}-\frac{1}{2}\right)}{\Gamma\left(\frac{1}{q-1}\right)}}=\frac{\Gamma\left(\frac{1}{q-1}\right)\left[\left(q-1\right)a\right]^{1/2}}{\sqrt{\pi}\Gamma\left(\frac{1}{q-1}-\frac{1}{2}\right)},\label{eq:normalization-factor-d-1-qmaior1}
\end{equation}
once $\Gamma\left(\frac{1}{2}\right)=\sqrt{\pi}$ (see \cite{Bellandi-Filho1986},
page 37).

\subsection{Case $d=1$ and $q<1$}

In this case, expression (\ref{eq:integral-normalization-factor-1-d})
is well defined only if:

\[
1+\left(q-1\right)ay\geq0,
\]
which implies:

\[
|y|\leq\frac{1}{\sqrt{\left(1-q\right)a}}=\left(\left(1-q\right)a\right)^{-1/2}.
\]

Hence, equation (\ref{eq:integral-normalization-factor-1-d}) becomes:

\[
\int_{\mathbb{R}}\left[1+\left(q-1\right)\left(ay^{2}\right)\right]^{\frac{1}{1-q}}dy=\int_{-\left(\left(1-q\right)a\right)^{-1/2}}^{+\left(\left(1-q\right)a\right)^{-1/2}}\left[1+\left(q-1\right)\left(ay^{2}\right)\right]^{\frac{1}{1-q}}dy.
\]

By using the variable change:

\[
\widetilde{y}=\left(\left(1-q\right)a\right)^{1/2}y,
\]
we get:

\begin{equation}
\int_{-1}^{+1}\left[1-\widetilde{y}^{2}\right]^{\frac{1}{1-q}}\frac{d\widetilde{y}}{\left(\left(1-q\right)a\right)^{1/2}}.\label{eq:normalization-1m1-changed}
\end{equation}

From equation 3.251-1 of reference \cite{gradshteyn1981}, we see:

\begin{equation}
\int_{0}^{1}x^{\mu-1}\left(1-x^{\lambda}\right)^{\nu-1}dx=\frac{1}{\lambda}B\left(\frac{\mu}{\lambda},\nu\right),\quad if\quad Re\left(\mu\right)>0,\; Re\left(\nu\right)>0,\;\lambda>0,\label{eq:formula-table-int-000}
\end{equation}
where $B$ is the Beta function given by expression (\ref{eq:def-function-beta}).

Therefore, by setting:

\[
\mu=1,\;\lambda=2,\;\nu=\frac{1}{1-q}+1,
\]
in expression (\ref{eq:formula-table-int-000}) we obtain:

\[
\int_{0}^{1}\left(1-x^{2}\right)^{\frac{1}{1-q}}dx=\frac{1}{2}B\left(\frac{1}{2},\frac{1}{1-q}+1\right).
\]

Hence, we can compute equation (\ref{eq:normalization-1m1-changed})
as:

\[
\frac{1}{\left(\left(1-q\right)a\right)^{1/2}}\int_{-1}^{+1}\left[1-\widetilde{y}^{2}\right]^{\frac{1}{1-q}}d\widetilde{y}
\]

\[
\frac{2}{\left(\left(1-q\right)a\right)^{1/2}}\int_{0}^{1}\left(1-x^{2}\right)^{\frac{1}{1-q}}dx=\frac{1}{\left(\left(1-q\right)a\right)^{1/2}}B\left(\frac{1}{2},\frac{1}{1-q}+1\right).
\]

By inserting this result and equation (\ref{eq:def-function-beta})
in expression (\ref{eq:normalization-factor-1}) and we obtain:

\[
C_{1,q}\left(\beta\right)=\frac{1}{\int_{\mathbb{R}}\exp_{q}\left(-\frac{\beta}{\sigma^{2}}y^{2}\right)dy}
\]

\[
=\frac{1}{\frac{1}{\left(\left(1-q\right)a\right)^{1/2}}B\left(\frac{1}{2},\frac{1}{1-q}+1\right)}
\]

\[
=\frac{1}{\frac{1}{\left(\left(1-q\right)a\right)^{1/2}}B\left(\frac{1}{2},\frac{1}{1-q}+1\right)}
\]

\[
=\frac{1}{\frac{1}{\left(\left(1-q\right)a\right)^{1/2}}\frac{\Gamma\left(\frac{1}{2}\right)\Gamma\left(\frac{1}{1-q}+1\right)}{\Gamma\left(\frac{1}{1-q}+\frac{3}{2}\right)}}
\]

\begin{equation}
=\frac{\Gamma\left(\frac{1}{1-q}+\frac{3}{2}\right)\left(\left(1-q\right)a\right)^{1/2}}{\Gamma\left(\frac{1}{2}\right)\Gamma\left(\frac{1}{1-q}+1\right)}.\label{eq:normalization-factor-d-1-qmenor1}
\end{equation}
for $q<1$.

\subsection{One-Dimensional $q$-Gaussian Expression\label{sub:C.3-One-Dimensional-Q-Gaussian-Expression}}

By inserting equations (\ref{eq:normalization-factor-d-1-qmaior1})
and (\ref{eq:normalization-factor-d-1-qmenor1}) in expression (\ref{eq:Q-Gaussian-d})
with $d=1$we get:

\begin{equation}
G_{1,q}\left(x,\sigma,\beta\right)=C_{1,q}\left(\sigma,\beta\right)\exp_{q}\left(-\frac{\beta}{\sigma^{2}}x^{2}\right),\label{eq:Q-Gaussian-d-1}
\end{equation}
where:

\[
C_{1,q}\left(\sigma,\beta\right)=\frac{\Gamma\left(\frac{1}{q-1}\right)\left[\left(q-1\right)\frac{\beta}{\sigma^{2}}\right]^{1/2}}{\sqrt{\pi}\Gamma\left(\frac{1}{q-1}-\frac{1}{2}\right)},\quad3>q>1,
\]

\[
C_{1,q}\left(\sigma,\beta\right)=\frac{\Gamma\left(\frac{1}{1-q}+\frac{3}{2}\right)\left(\left(1-q\right)\frac{\beta}{\sigma^{2}}\right)^{1/2}}{\sqrt{\pi}\Gamma\left(\frac{1}{1-q}+1\right)},\quad q<1,\quad|x|\leq\left(\left(1-q\right)\frac{\beta}{\sigma^{2}}\right)^{-1/2}.
\]

Consequently, using equation A7 of reference \cite{Borges2012}, which
states that:

\[
\lim_{q\rightarrow1^{+}}\frac{\Gamma\left(\frac{1}{q-1}-\alpha\right)}{\left(q-1\right)^{\alpha}\Gamma\left(\frac{1}{q-1}\right)}=1,\quad1<q<1+\frac{1}{\alpha},
\]

\[
\lim_{q\rightarrow1^{-}}\frac{\Gamma\left(\frac{1}{1-q}+1\right)}{\left(1-q\right)^{\alpha}\Gamma\left(\frac{1}{1-q}+\alpha+1\right)}=1,\quad q<1,
\]
we can show that:

\[
\lim_{q\rightarrow1}G_{1,q}\left(\mathbf{x},\sigma,\beta\right)=\frac{\sqrt{\beta}}{\sigma\sqrt{\pi}}\exp\left(-\frac{\beta}{\sigma^{2}}x^{2}\right).
\]

If $\beta=0.5$, we get:

\begin{equation}
\lim_{q\rightarrow1}G_{1,q}\left(\mathbf{x},\sigma,0.5\right)=\frac{1}{\sigma\sqrt{2\pi}}\exp\left(-\frac{x^{2}}{2\sigma^{2}}\right),\label{eq:traditional-gaussian}
\end{equation}
which is the traditional Gaussian.

\section{Appendix: Two-Dimensional $q$-Gaussian \label{sec:Appenix-D-Two-Dimensional} }

If $d=2$ then we can write $\mathbf{z}=\left(z_{1},z_{2}\right)$
and, using polar coordinates:

\begin{equation}
z_{1}=r\cos\left(\theta\right),\label{eq:polar00}
\end{equation}

\begin{equation}
z_{2}=r\sin\left(\theta\right),\label{eq:polar01}
\end{equation}

\[
det\left(J\right)=r,
\]
we can write the integral in expression (\ref{eq:normalization-factor})
as:

\[
\int_{\mathbb{R}^{2}}\exp_{q}\left(-\beta|\mathbf{z|}^{2}\right)d\mathbf{z}=\int_{0}^{2\pi}\int_{0}^{+\infty}r\exp_{q}\left(-\beta r^{2}\right)drd\theta.
\]

In the above development we are supposing $\beta>0$.

\subsection{Case $q>1$}

Then:

\[
\int_{0}^{2\pi}\int_{0}^{+\infty}r\exp_{q}\left(-\beta r^{2}\right)drd\theta=2\pi\int_{0}^{+\infty}r\exp_{q}\left(-\beta r^{2}\right)dr
\]

\[
=2\pi\int_{0}^{+\infty}r\left[1+\left(1-q\right)\left(-\beta r^{2}\right)\right]^{\frac{1}{1-q}}dr
\]

\begin{equation}
=2\pi\int_{0}^{+\infty}r\left[1-\left(1-q\right)\beta r^{2}\right]^{\frac{1}{1-q}}dr.\label{eq:double-integral01}
\end{equation}

Variable change:

\begin{equation}
u=1-\left(1-q\right)\beta r^{2},\label{eq:variable-change00}
\end{equation}

\begin{equation}
du=-2\left(1-q\right)\beta rdr=2\left(q-1\right)\beta rdr.\label{eq:variable-change01}
\end{equation}

Besides, once $q>1$ and $\beta>0$, we have

\[
r\rightarrow+\infty\Longrightarrow u\rightarrow+\infty,\quad and\quad u\left(0\right)=1.
\]

Therefore, using the variable change defined by expressions (\ref{eq:variable-change00})-(\ref{eq:variable-change01})
we obtain:

\[
2\pi\int_{0}^{+\infty}r\left[1-\left(1-q\right)\beta r^{2}\right]^{\frac{1}{1-q}}dr=\frac{2\pi}{2\left(q-1\right)\beta}\int_{0}^{+\infty}\left[1-\left(1-q\right)\beta r^{2}\right]^{\frac{1}{1-q}}\left(2\left(q-1\right)\beta rdr\right)
\]

\[
=\frac{2\pi}{2\left(q-1\right)\beta}\int_{1}^{+\infty}u^{\frac{1}{1-q}}du=\frac{2\pi}{2\left(q-1\right)\beta}\left[\frac{u^{\frac{1}{1-q}+1}}{\frac{1}{1-q}+1}\right]_{1}^{+\infty}=\frac{2\pi}{2\left(q-1\right)\beta}\left[\frac{u^{\frac{2-q}{1-q}}}{\frac{2-q}{1-q}}\right]_{1}^{+\infty}.
\]

The above integral converges only if:

\[
\frac{1}{1-q}+1=\frac{2-q}{1-q}<0\quad\Longrightarrow\quad q<2.
\]

Once we are considering $q>1$, we get that, if $1<q<2$ then:

\begin{equation}
2\pi\int_{0}^{+\infty}r\left[1-\left(1-q\right)\beta r^{2}\right]^{\frac{1}{1-q}}dr=\frac{2\pi}{2\left(q-1\right)\beta}\left(-\frac{1}{\frac{2-q}{1-q}}\right)=\frac{2\pi}{2\left(q-1\right)\beta}\left(\frac{q-1}{2-q}\right)=\frac{\pi}{\beta\left(2-q\right)}.\label{eq:1-menor-q-menor-2}
\end{equation}

\subsection{Case $q<1$\label{sub:D.2-Case-q-menor-1}}

In this case, in order to get a real value in the integral given by
expression (\ref{eq:double-integral01}) we need:

\begin{equation}
1-\left(1-q\right)\beta r^{2}>0\quad\Longrightarrow\quad0<r<\frac{1}{\sqrt{\beta\left(1-q\right)}}.\label{eq:restriction00}
\end{equation}

Then, with the constraints $q<1$ and $0<r<\left(\beta\left(1-q\right)\right)^{-1/2}$,
we can use the same variable change as before:

\[
u=u\left(r\right)=1+\left(q-1\right)\beta r^{2}>0.
\]

By inserting it in the integral (\ref{eq:double-integral01}) we obtain:

\[
2\pi\int_{0}^{+\frac{1}{\sqrt{1-q}}}r\left[1-\left(1-q\right)r^{2}\right]^{\frac{1}{1-q}}dr=\left(\frac{2\pi}{2\left(q-1\right)\beta}\right)\left[\frac{u^{\frac{2-q}{1-q}}}{\frac{2-q}{1-q}}\right]_{1}^{0}
\]

\[
=\left(\frac{2\pi}{2\left(q-1\right)\beta}\right)\left[0-\frac{\left(1-q\right)}{\left(2-q\right)}\right]=\frac{\pi}{\beta\left(2-q\right)}.
\]

\subsection{$q$-Gaussian 2D: Putting All Together}

Therefore, for $d=2$, expression (\ref{eq:normalization-factor-Q-Gaussian})
gives:

\begin{equation}
C_{2,q}\left(\Sigma,\beta\right)=\left(\int_{\mathbb{R}^{2}}\exp_{q}\left(-\beta\mathbf{y^{T}}D^{-1}\mathbf{y}\right)d\mathbf{y}\right)^{-1}=\left(\int_{\mathbb{R}^{2}}\exp_{q}\left(-\beta|\mathbf{z|}^{2}\right)\sqrt{|\Sigma|}d\mathbf{z}\right)^{-1}
\end{equation}

\begin{equation}
=\frac{1}{\sqrt{|\Sigma|}\int_{\mathbb{R}^{2}}\exp_{q}\left(-\beta|\mathbf{z|}^{2}\right)d\mathbf{z}}\label{eq:normalization-final00}
\end{equation}

Then, for $\beta>0$, expression (\ref{eq:1-menor-q-menor-2}) renders:

\begin{equation}
\int_{\mathbb{R}^{2}}\exp_{q}\left(-\beta|\mathbf{z|}^{2}\right)d\mathbf{z}=\frac{\pi}{\beta\left(2-q\right)},\quad if\quad1<q<2,\label{eq:integral00}
\end{equation}

If $q<1$, we have the restriction:

\begin{equation}
\Omega=\left\{ r\in\mathbb{R};\,0<r<\frac{1}{\sqrt{\beta\left(1-q\right)}}\right\} ,\label{eq:constraint00}
\end{equation}
due to expression (\ref{eq:restriction00}). Therefore, according
to section \ref{sub:D.2-Case-q-menor-1}, the integral in equation
(\ref{eq:normalization-final00}) becomes:

\[
\int_{\mathbb{R}^{2}}\exp_{q}\left(-\beta|\mathbf{z|}^{2}\right)d\mathbf{z}=\int_{0}^{2\pi}\int_{\Omega}r\exp_{q}\left(-\beta r^{2}\right)drd\theta=\frac{\pi}{\beta\left(2-q\right)}.
\]

By assembling all the above results we obtain that, for $d=2$, expression
(\ref{eq:Q-Gaussian-d-2}) becomes:

\begin{equation}
G_{2,q}\left(\mathbf{x},\Sigma,\beta\right)=\frac{\beta\left(2-q\right)}{\pi\sqrt{|\Sigma|}}\left[1+\left(1-q\right)\left(-\beta\mathbf{x^{T}}\Sigma^{-1}\mathbf{x}\right)\right]^{^{\frac{1}{1-q}}},\quad1<q<2,\label{eq:final-2D-gaussian00}
\end{equation}

\begin{equation}
G_{2,q}\left(\mathbf{x},\Sigma,\beta\right)=\frac{\beta\left(2-q\right)}{\pi\sqrt{|\Sigma|}}\left[1+\left(1-q\right)\left(-\beta\mathbf{x^{T}}\Sigma^{-1}\mathbf{x}\right)\right]^{^{\frac{1}{1-q}}},\quad q<1,\label{eq:final-2D-gaussian01}
\end{equation}
subject to the constraint:
\[
\quad0<\left(\mathbf{x}^{T}\Sigma^{-1}\mathbf{x}\right)^{1/2}<\frac{1}{\sqrt{\beta\left(1-q\right)}}.
\]

Consequently:

\[
\lim_{q\rightarrow1}G_{2,q}\left(\mathbf{x},\Sigma,\beta\right)=\left[\lim_{q\rightarrow1}C_{2,q}\left(\mathbf{x},\beta\right)\right]\left[\lim_{q\rightarrow1}\exp_{q}\left(-\beta\mathbf{x^{T}}\Sigma^{-1}\mathbf{x}\right)\right]=\frac{\beta}{\pi\sqrt{|\Sigma|}}\exp\left(-\beta\mathbf{x^{T}}\Sigma^{-1}\mathbf{x}\right).
\]

So, by setting $\beta=1/2$, we get:

\[
\lim_{q\rightarrow1}G_{2,q}\left(\mathbf{x,}\Sigma,\frac{1}{2}\right)=\left[\lim_{q\rightarrow1}C_{2,q}\left(\mathbf{x,}\frac{1}{2}\right)\right]\left[\lim_{q\rightarrow1}\exp_{q}\left(-\frac{1}{2}\mathbf{x^{T}}\Sigma^{-1}\mathbf{x}\right)\right]=\frac{1}{2\pi\sqrt{|\Sigma|}}\exp\left(-\frac{1}{2}\mathbf{x^{T}}\Sigma^{-1}\mathbf{x}\right),
\]
which is the traditional two dimensional Gaussian.

\section{Appendix: Fourier Transform of $q$-Exponential\label{sec:Appendix-E-Fourier}}

Let:

\[
\mathcal{F}\left(\exp_{q}\left(-ax^{2}\right);y\right)=\int_{-\infty}^{+\infty}\exp\left(-2j\pi xy\right)\left[1-\left(1-q\right)ax^{2}\right]^{\frac{1}{1-q}}dx
\]

\begin{equation}
=\int_{-\infty}^{+\infty}\exp\left(-2j\pi xy\right)\left[1+\left(q-1\right)ax^{2}\right]^{\frac{1}{1-q}}dx.\label{eq:fourier-transform-q00}
\end{equation}

Variable change:

\begin{equation}
\tilde{x}=\left[\left(q-1\right)a\right]^{1/2}x;\quad\tilde{y}=\left[\left(q-1\right)a\right]^{-1/2}y.\label{eq:variable-change000}
\end{equation}

Then:

\begin{equation}
\mathcal{F}\left(\exp_{q}\left(-ax^{2}\right);y\right)=\left[\left(q-1\right)a\right]^{-1/2}\int_{-\infty}^{+\infty}\exp\left(-2j\pi\tilde{x}\tilde{y}\right)\left[1+\tilde{x}^{2}\right]^{\frac{1}{1-q}}d\tilde{x}.\label{eq:q-fourier-expression00}
\end{equation}

We are going to use the fact that:

\[
\int_{-\infty}^{+\infty}\left(\beta+jx\right)^{-2\mu}\left(\gamma-jx\right)^{-2\nu}e^{-jpx}dx
\]

\[
=-2\pi\left(\beta+\gamma\right)^{-\mu-\nu}\frac{p^{\mu+\nu-1}}{\Gamma\left(2\nu\right)}\exp\left(\frac{\gamma-\beta}{2}p\right)W_{\mu-\nu,\frac{1}{2}-\nu-\mu}\left(\beta p+\gamma p\right),\quad p>0,
\]

\[
=2\pi\left(\beta+\gamma\right)^{-\mu-\nu}\frac{\left(-p\right)^{\mu+\nu-1}}{\Gamma\left(2\mu\right)}\exp\left(\frac{\beta-\gamma}{2}p\right)W_{\mu-\nu,\frac{1}{2}-\nu-\mu}\left(-\beta p-\gamma p\right),\quad p<0,
\]
if $Re\left(\beta\right)>0$, $Re\left(\gamma\right)>0$ and $Re\left(\mu+\nu\right)>1/2$,
where $W$ denotes the Whittaker functions (see expressions 9-220
of reference \cite{gradshteyn1981}).

Therefore, if $\beta=\gamma=1$ and $\mu=\nu$ we have:

\[
\left(\beta+jx\right)^{-2\mu}\left(\gamma-jx\right)^{-2\nu}=\left(1+jx\right)^{-2\mu}\left(1-jx\right)^{-2\mu}=\left(1+x^{2}\right)^{-2\mu},
\]
and:

\[
\int_{-\infty}^{+\infty}\left(1+x^{2}\right)^{-2\mu}e^{-jpx}dx
\]

\begin{equation}
=-2\pi\left(2\right)^{-2\mu}\frac{p^{2\mu-1}}{\Gamma\left(2\mu\right)}W_{0,\frac{1}{2}-2\mu}\left(2p\right),\quad p>0,\label{eq:q-fourier00}
\end{equation}

\begin{equation}
=2\pi\left(2\right)^{-2\mu}\frac{\left(-p\right)^{2\mu-1}}{\Gamma\left(2\mu\right)}W_{0,\frac{1}{2}-2\mu}\left(-2p\right),\quad p<0,\label{eq:q-fourier01}
\end{equation}
where $W_{0,\frac{1}{2}-2\mu}$ is defined by expression (\ref{eq:Whittaker00}).

So, we can put expressions (\ref{eq:q-fourier00})-(\ref{eq:q-fourier01})
together to obtain:

\begin{equation}
\int_{-\infty}^{+\infty}\left(1+x^{2}\right)^{-2\mu}e^{-jpx}dx=-sign\left(p\right)2\pi\left(2^{-2\mu}\right)\frac{|p|^{2\mu-1}}{\Gamma\left(2\mu\right)}W_{0,\frac{1}{2}-2\mu}\left(2|p|\right)
\end{equation}

\begin{equation}
=-sign\left(p\right)2\pi\frac{1}{2}\frac{1}{2^{2\mu-1}}\frac{|p|^{2\mu-1}}{\Gamma\left(2\mu\right)}W_{0,\frac{1}{2}-2\mu}\left(2|p|\right)
\end{equation}

\begin{equation}
=-sign\left(p\right)\pi\frac{1}{\Gamma\left(2\mu\right)}\left(\frac{|p|}{2}\right)^{2\mu-1}W_{0,\frac{1}{2}-2\mu}\left(2|p|\right),\label{eq:q-fourier-geral}
\end{equation}
if $Re\left(2\mu\right)>1/2$.

Now, let us return to expression (\ref{eq:q-fourier-expression00}).
By setting: $p=2\pi\tilde{y}$ and $2\mu=\left(q-1\right)^{-1}$,
with the constraint:

\begin{equation}
Re\left(2\mu\right)>1/2\quad\Longrightarrow\quad\left(q-1\right)^{-1}>\frac{1}{2},\label{eq:constraint-real}
\end{equation}
we get the following cases.

\subsection{Case $q>1$}
Due to the restriction (\ref{eq:constraint-real}), we have $1<q<3$.
By inserting equation (\ref{eq:q-fourier-geral}) into expression
(\ref{eq:q-fourier-expression00}) we obtain:

\[
\mathcal{F}\left(\exp_{q}\left(-ax^{2}\right);y\right)=\left[\left(q-1\right)a\right]^{-1/2}\int_{-\infty}^{+\infty}\exp\left(-2j\pi\tilde{x}\tilde{y}\right)\left[1+\tilde{x}^{2}\right]^{\frac{1}{1-q}}d\tilde{x}
\]

\[
=\left[\left(q-1\right)a\right]^{-1/2}\left(-sign\left(2\pi\tilde{y}\right)\pi\frac{1}{\Gamma\left(\frac{1}{q-1}\right)}\left(\frac{|2\pi\tilde{y}|}{2}\right)^{2\mu-1}W_{0,\frac{1}{2}-\frac{1}{q-1}}\left(2|2\pi\tilde{y}|\right)\right)
\]

\[
=\left[\left(q-1\right)a\right]^{-1/2}\left(-sign\left(2\pi\left[\left(q-1\right)a\right]^{-1/2}y\right)2\pi\left(2^{\frac{1}{1-q}}\right)\frac{|2\pi\left[\left(q-1\right)a\right]^{-1/2}y|^{\frac{1}{q-1}-1}}{\Gamma\left(\frac{1}{q-1}\right)}\right)\times
\]

\begin{equation}
W_{0,\frac{1}{2}-\frac{1}{q-1}}\left(2|2\pi\left[\left(q-1\right)a\right]^{-1/2}y|\right),\label{eq:fourier-1d-transform-q-gauss00}
\end{equation}
where the function $W_{0,\frac{1}{2}-\frac{1}{q-1}}$ is defined through
equation (\ref{eq:Whittaker00}). Using the fact that:

\[
W_{0,\mu}\left(z\right)=W_{0,-\mu}\left(z\right),
\]
we can write:

\[
|\mathcal{F}\left(\exp_{q}\left(-ax^{2}\right);y\right)|
\]

\[
=abs\left[\left[\left(q-1\right)a\right]^{-1/2}\left(2\pi\left(2^{\frac{1}{1-q}}\right)\frac{|2\pi\left[\left(q-1\right)a\right]^{-1/2}y|^{\frac{1}{q-1}-1}}{\Gamma\left(\frac{1}{q-1}\right)}\right)\right]\times
\]

\[
abs\left(\frac{2|2\pi\left[\left(q-1\right)a\right]^{-1/2}y|}{\pi}\right)^{1/2}\times
\]

\[
abs\left[\frac{i\pi}{2}\exp\left(\frac{i\pi}{2}\left(-\frac{1}{2}-\frac{1}{1-q}\right)\right)H_{-\frac{1}{2}-\frac{1}{1-q}}^{1}\left(i\frac{2|2\pi\left[\left(q-1\right)a\right]^{-1/2}y|}{2}\right)\right].
\]

\subsection{Case $q<1$}

In this case, expression (\ref{eq:fourier-transform-q00}) is well
defined only if:

\[
1+\left(q-1\right)ax^{2}\geq0,
\]
which implies:

\[
|x|\leq\frac{1}{\sqrt{\left(1-q\right)a}}=\left(\left(1-q\right)a\right)^{-1/2}.
\]

Hence, equation (\ref{eq:fourier-transform-q00}) becomes:

\begin{equation}
\int_{-\left(\left(1-q\right)a\right)^{-1/2}}^{+\left(\left(1-q\right)a\right)^{-1/2}}\exp\left(-2j\pi xy\right)\left[1+\left(q-1\right)ax^{2}\right]^{\frac{1}{1-q}}dx.\label{eq:fourier-transform-qm1}
\end{equation}

By using the variable change:

\[
\widetilde{x}=\left(\left(1-q\right)a\right)^{1/2}x,
\]

\[
\widetilde{y}=\left(\left(1-q\right)a\right)^{-1/2}y,
\]
we can rewrite equation (\ref{eq:fourier-transform-qm1}) as:

\begin{equation}
\int_{-1}^{+1}\exp\left(-2j\pi\widetilde{x}\widetilde{y}\right)\left[1-\widetilde{x}^{2}\right]^{\frac{1}{1-q}}\frac{d\widetilde{x}}{\left(\left(1-q\right)a\right)^{1/2}}.\label{eq:fourier-transform-1m1-changed}
\end{equation}

However, it is known that (equation 3.387-2 of \cite{gradshteyn1981}):

\begin{equation}
\int_{-1}^{+1}\left(1-x^{2}\right)^{\nu-1}\exp\left(j\mu x\right)dx=\sqrt{\pi}\left(\frac{2}{\mu}\right)^{\nu-\frac{1}{2}}\Gamma\left(\nu\right)\mathbb{J}_{\nu-\frac{1}{2}}\left(\mu\right),\label{eq:integral-from-table}
\end{equation}
if $Re\left(\nu\right)>0$.

Therefore, if we set:

\[
\nu=\frac{1}{1-q}+1,
\]

\[
\mu=-2\pi\widetilde{y},
\]

\[
x=\tilde{x},
\]
in expression (\ref{eq:integral-from-table}), we obtain $Re\left(\nu\right)>0$,
and:

\[
\int_{-1}^{+1}\left(1-\widetilde{x}^{2}\right)^{\frac{1}{1-q}}\exp\left(-2j\pi\widetilde{x}\widetilde{y}\right)d\widetilde{x}
\]

\[
=\sqrt{\pi}\left(-\frac{1}{\pi\widetilde{y}}\right)^{\frac{1}{1-q}+\frac{1}{2}}\Gamma\left(\frac{1}{1-q}+1\right)\mathbb{J}_{\frac{1}{1-q}+\frac{1}{2}}\left(-2\pi\widetilde{y}\right),\; if\; y\neq0,
\]
where $\mathbb{J}$ denotes the Bessel functions (see section \ref{sec:Appendix-F-Special-Functions}).

Consequently, we finally have:

\[
\mathcal{F}\left(\exp_{q}\left(-ax^{2}\right);y\right)=\int_{-\left(\left(1-q\right)a\right)^{-1/2}}^{+\left(\left(1-q\right)a\right)^{-1/2}}\exp\left(-2j\pi xy\right)\left[1+\left(q-1\right)ax^{2}\right]^{\frac{1}{1-q}}dx
\]

\[
=\frac{1}{\left(\left(1-q\right)a\right)^{1/2}}\int_{-1}^{+1}\exp\left(-2j\pi\widetilde{x}\widetilde{y}\right)\left[1-\widetilde{x}^{2}\right]^{\frac{1}{1-q}}d\widetilde{x}
\]

\[
=\frac{1}{\left(\left(1-q\right)a\right)^{1/2}}\sqrt{\pi}\left(-\frac{1}{\pi\widetilde{y}}\right)^{\frac{1}{1-q}+\frac{1}{2}}\Gamma\left(\frac{1}{1-q}+1\right)\mathbb{J}_{\frac{1}{1-q}+\frac{1}{2}}\left(-2\pi\widetilde{y}\right)
\]

\[
=\frac{\sqrt{\pi}}{\left(\left(1-q\right)a\right)^{1/2}}\left(-\frac{1}{\pi\left[\left(\left(1-q\right)a\right)^{-1/2}y\right]}\right)^{\frac{1}{1-q}+\frac{1}{2}}\Gamma\left(\frac{1}{1-q}+1\right)\mathbb{J}_{\frac{1}{1-q}+\frac{1}{2}}\left(-2\pi\left[\left(\left(1-q\right)a\right)^{-1/2}y\right]\right)
\]

\begin{equation}
=\frac{\sqrt{\pi}}{\left(\left(1-q\right)a\right)^{1/2}}\Gamma\left(\frac{1}{1-q}+1\right)\left(-\frac{\left(\left(1-q\right)a\right)^{1/2}}{\pi y}\right)^{\frac{1}{1-q}+\frac{1}{2}}\mathbb{J}_{\frac{1}{1-q}+\frac{1}{2}}\left(-\frac{2\pi y}{\left(\left(1-q\right)a\right)^{1/2}}\right),\label{eq:fourier-1d-transform-q-gauss01}
\end{equation}
if $q<1$ and $y\neq0$.

If $y=0$ in expression (\ref{eq:fourier-transform-qm1}) we get:

\[
\int_{-\left(\left(1-q\right)a\right)^{-1/2}}^{+\left(\left(1-q\right)a\right)^{-1/2}}\left[1+\left(q-1\right)ax^{2}\right]^{\frac{1}{1-q}}dx
\]

\begin{equation}
=\int_{-1}^{+1}\left[1-\widetilde{x}^{2}\right]^{\frac{1}{1-q}}\frac{d\widetilde{x}}{\left(\left(1-q\right)a\right)^{1/2}}.\label{eq:transf-qgauss-y0}
\end{equation}

However, for expression 3.214, reference \cite{gradshteyn1981}, we
have:

\[
\int_{0}^{1}\left[\left(1+x\right)^{\mu-1}\left(1-x\right)^{\nu-1}+\left(1+x\right)^{\nu-1}\left(1-x\right)^{\mu-1}\right]dx=2^{\mu+\nu-1}B\left(\mu,\nu\right),\; Re\left(\mu\right)>0,Re\left(\nu\right)>0.
\]

Therefore, if :

\[
\mu=\nu=\frac{1}{\left(1-q\right)}+1,
\]
we satisfy $Re\left(\mu\right)>0,Re\left(\nu\right)>0$ and:

\[
2\int_{0}^{1}\left[\left(1+x\right)^{\frac{1}{\left(1-q\right)}}\left(1-x\right)^{\frac{1}{\left(1-q\right)}}\right]dx=2\int_{0}^{1}\left(1-x^{2}\right)^{\frac{1}{\left(1-q\right)}}dx=2^{\frac{2}{1-q}+1}B\left(\frac{1}{\left(1-q\right)}+1,\frac{1}{\left(1-q\right)}+1\right).
\]

So, expression (\ref{eq:transf-qgauss-y0}) can be computed by:

\[
\mathcal{F}\left(\exp_{q}\left(-ax^{2}\right);0\right)
\]

\[
=\frac{1}{\left(\left(1-q\right)a\right)^{1/2}}\int_{-1}^{+1}\left[1-\widetilde{x}^{2}\right]^{\frac{1}{1-q}}d\widetilde{x}=\frac{2^{\frac{2}{1-q}+1}}{\left(\left(1-q\right)a\right)^{1/2}}B\left(\frac{1}{\left(1-q\right)}+1,\frac{1}{\left(1-q\right)}+1\right)
\]

\[
=\frac{2^{\frac{2}{1-q}+1}}{\left(\left(1-q\right)a\right)^{1/2}}\frac{\Gamma\left(\frac{1}{\left(1-q\right)}+1\right)\Gamma\left(\frac{1}{\left(1-q\right)}+1\right)}{\Gamma\left(\frac{2}{1-q}+2\right)}.
\]

\section{Appendix: Special Functions\label{sec:Appendix-F-Special-Functions}}
\begin{itemize}
\item Gamma Function: Defined by Euler in the form of the product:
\end{itemize}
\begin{equation}
\Gamma\left(z\right)=\frac{1}{z}\prod_{n=1}^{+\infty}\left\{ \left(1+\frac{1}{n}\right)^{z}\left(1+\frac{z}{n}\right)^{-1}\right\} ,\label{eq:gamma-function-definition-Euler}
\end{equation}
which is not valid for $z\in\mathbb{Z}_{-}$.
\begin{itemize}
\item Beta Function: It is defined by:
\end{itemize}
\begin{equation}
B\left(x,y\right)=\frac{\Gamma\left(x\right)\Gamma\left(y\right)}{\Gamma\left(x+y\right)},\label{eq:def-function-beta}
\end{equation}
where $\Gamma$ is the Gamma function given by expression (\ref{eq:gamma-function-definition-Euler}).
We shall notice that the above expression is not defined if $x,y\in\mathbb{Z}_{-}$
\begin{itemize}
\item Whittaker Functions $W_{\lambda,\mu}\left(z\right)$: The general
case is given by expressions 9.220 of reference \cite{gradshteyn1981}.
In this work, we apply a special kind obtained by setting $\lambda=0$.
In this case, we have:
\end{itemize}
\begin{equation}
W_{0,\mu}\left(z\right)=\sqrt{\frac{z}{\pi}}K_{\mu}\left(\frac{z}{2}\right),\label{eq:Whittaker00}
\end{equation}
where $K_{\mu}$ are the modified Bessel functions (equation 8.407,
reference \cite{gradshteyn1981}):

\begin{equation}
K_{\mu}\left(\frac{z}{2}\right)=\frac{i\pi}{2}\exp\left(\frac{i\pi\mu}{2}\right)H_{\mu}^{1}\left(i\frac{z}{2}\right),\label{eq:Whittaker01}
\end{equation}
and (equations 8.405 and 8.403, reference \cite{gradshteyn1981}):

\begin{equation}
H_{\mu}^{1}\left(i\frac{z}{2}\right)=\mathbb{J}_{\mu}\left(i\frac{z}{2}\right)+i\frac{\left[\mathbb{J}_{\mu}\left(i\frac{z}{2}\right)\cos\left(\pi\mu\right)-\mathbb{J}_{-\mu}\left(i\frac{z}{2}\right)\right]}{\sin\left(\pi\mu\right)},\label{eq:degenerate-hypergeometric}
\end{equation}
with $\mathbb{J}_{\mu}$ being the Bessel functions (see bellow).
These expressions can be computed only if: $\mu\notin\mathbb{Z}_{-}$
and $|arg\left(z\right)|<\pi$.
\begin{itemize}
\item Bessel Functions: They are defined by \cite{Bellandi-Filho1986}:
\end{itemize}
\begin{equation}
\mathbb{J}_{v}\left(z\right)=\left(\frac{z}{2}\right)^{\nu}\sum_{k=0}^{+\infty}\frac{\left(-\frac{z^{2}}{4}\right)^{k}}{k!\Gamma\left(\nu+k+1\right)},\label{eq:Bessel-functions}
\end{equation}
where $\nu$ is real and $z$ can be complex. Considering the restriction
for gamma function, we should not have $\left(\nu+k+1\right)\in\mathbb{Z}_{-}$.

\bibliographystyle{sbc}
\bibliography{holography}

\end{document}